\documentclass[journal,twoside,web]{ieeecolor}
\usepackage{generic}
\usepackage{cite}
\usepackage{amsmath,amssymb,amsfonts}
\usepackage{algorithmic}
\usepackage{graphicx}
\usepackage{textcomp}
\def\BibTeX{{\rm B\kern-.05em{\sc i\kern-.025em b}\kern-.08em
    T\kern-.1667em\lower.7ex\hbox{E}\kern-.125emX}}
\usepackage{microtype}
\usepackage{graphicx}
\usepackage[colorlinks, citecolor=blue!90]{hyperref}
\usepackage{booktabs} % for professional tables
\usepackage[most]{tcolorbox}
\usepackage{bbm}
\usepackage{multirow}
\usepackage{makecell}
\usepackage{float}
\usepackage{bm}
\usepackage{nicefrac}
\usepackage{caption}
\usepackage{todonotes}
\usepackage{subfig}
\usepackage{makecell}

\newcommand{\method}{\textit{Deep Survival Machines}}
\newcommand{\dsm}{DSM}

\begin{document}
\title{\method{}:\\ Fully Parametric Survival Regression\\ and Representation Learning\\ for Censored Data with Competing Risks}
\author{Chirag Nagpal, Xinyu Rachel Li, and Artur Dubrawski 
\thanks{All authors are affliated with the Auton Lab, School of Computer Science, Carnegie Mellon University
(e-mail: \{chiragn, xinyul2, awd \}@cs.cmu.edu). Chirag Nagpal and Xinyu Li are equal first authors.}}
\maketitle

\begin{abstract}
We describe a new approach to estimating relative risks in time-to-event prediction problems with censored data in a fully parametric manner. Our approach does not require making strong assumptions of constant proportional hazard of the underlying survival distribution, as  required by the Cox-proportional hazard model. By jointly learning deep nonlinear representations of the input covariates, we demonstrate the benefits of our approach when used to estimate survival risks through extensive experimentation on multiple real world datasets with different levels of censoring. We further demonstrate advantages of our model in the competing risks scenario. To the best of our knowledge, this is the first work involving fully parametric estimation of survival times with competing risks in the presence of censoring.
\end{abstract}


%{\small\bfseries\color{subsectioncolor}\textit{\textsf{Software Package}}---}\\
%\centerline{\textbf{\href{http://autonlab.github.io/DeepSurvivalMachines}{\texttt{autonlab.github.io/DeepSurvivalMachines}}}}

% \textit{\textbf{\textsf{}}}--

\section{Introduction}
\label{sec:introduction}
\IEEEPARstart{S}{urvival} regression is a field of statistics and machine learning that deals with the estimation of a survival function representing the probability of an event of interest, typically a failure, to occur beyond a certain time in the future. 
Survival regression models time-to-event by estimating the survival function, $\mathbb{S}(\cdot|X)\triangleq\mathbb{P}(T>t|X) $, conditional on $X$, the input covariates. Examples include estimating the survival times of patients after certain treatment using clinical variables, or predicting the failure times of machines using their usage histories, etc. Survival regression differs from standard regression due to censoring of data, i.e.\ observation of some subjects stops before occurrence of an event of interest. 
In practical settings, there might be multiple different events that may lead to failure, and this generalized setting is known as the \textit{competing risks} scenario.

Classical statistical machine learning techniques for survival regression rely on non-parametric or semi-parametric methods for survival function estimation, primarily because they make working with censored data relatively straightforward. However, non-parametric methods may suffer from curse of dimensionality, and semi-parametric approaches usually depend on strong modelling assumptions. In particular, the prevailing assumption of constant proportional hazard over lifetime as proposed by \cite{cox1972regression} in the Proportional Hazards model, is very likely to be unrealistic in many practical scenarios encountered in healthcare, predictive maintenance, econometrics, or operations research. % also known as \textit{`Coxian'} assumption
This and similar assumptions have recently attracted much controversy. 

In this paper, we propose \method{}, a novel approach to estimate time-to-event in the presence of censoring. By leveraging a hierarchical graphical model parameterized by neural networks, we learn distributional representations of the input covariates and mitigate existing challenges in survival regression. 

Our main contributions can be summarized as follows:
\begin{enumerate}
    \item Our approach estimates the conditional survival function $\mathbb{S}(\cdot|X)$ as a mixture of individual parametric survival distributions.
    \item We do not make strong assumptions of proportional hazards and enable learning with time-varying risks.
    % \item while allowing for estimation of the conditional survival function as a mixture of individual parametric survival distributions.
    \item Our approach allows for learning of rich distributed representations of the input covariates, helping knowledge transfer across multiple competing risks. 
    
    % \item And finally, allows for one to learn rich distributional representations of the input data, that can be leveraged in \textit{competing risks} scenarios.
\end{enumerate}

Through extensive experimentation on multiple datasets, we demonstrate advantages of our approach in both the single event and \textit{competing risks} scenarios as compared to classic survival analysis techniques as well as more modern competitive baselines. Software code implementing Deep Survival Machines is open-sourced and publicly available as a \texttt{python} package.\footnote{\normalsize  \href{http://autonlab.github.io/DeepSurvivalMachines}{\normalsize{\texttt{autonlab.github.io/DeepSurvivalMachines}}}}

%\textcolor{}{The python implementation of our approach, \method{}, is publicly available.}

\section{Related Work}

Time-to-event modeling in the presence of censoring is an important statistical estimation problem with applications in multiple domains including bio-statistics and health informatics~\cite{zhu2016deep, kim2019deep}, actuarial sciences~\cite{czado2002application} and econometrics~\cite{stepanova2002survival, bosco2006factors, jones2002econometric}. 
Time-to-event regression has also been increasingly prevalent in engineering, including applications to support predictive maintenance of equipment~\cite{wang2017predictive, hochstein2013survival} and in cyber-physical systems. \cite{mishra2017bhm} proposed a Bayesian hierarchical model to predict the end of life (EoL) and end of discharge (EoD) of Li-ion batteries using variable load profiles and discharge data, and \cite{Ghasemi2007} used a proportional hazards model to track the degradation of the system when planning for the optimal maintenance strategy.

% survival regression
The Cox proportional hazards regression model (CPH) is a popular choice for modeling the distribution of time-to-event. In this model, the estimator of the survival function conditional on $X$, $\mathbb{S}(\cdot|X)\triangleq\mathbb{P}(T>t|X) $, is assumed to have constant proportional hazard. Thus, the relative proportional hazard between individuals is constant across time. Another way of stating this assumption is that if an individual is at a higher risk of death at a certain time as compared to another individual, then the relative risk associated with the individual would be higher at any point along the lifetime of this individual. This is a very strong assumption which may not hold in many practical scenarios when the risks change over time.

Significant volume of recent research is focused on improving the CPH model. \cite{pmlr-v52-kraisangka16, kraisangka_druzdzel_2018} combined Bayesian networks with the CPH model to improve both the model interpretability and predictive power. Researchers have also tried to incorporate structural sparsity, regularization, as well as active and multitask learning when available data is scarce \cite{vinzamuri2014active,vinzamuri2013cox,li2016multi}. Other efforts have involved incorporating non-linear interactions between the covariates in the original model. \cite{rosen1999mixtures} proposed using a mixture of linear experts for the original Cox model. \cite{nagpal2019nonlinear} recently improved that approach with an objective based on variational inference and demonstrated state-of-the-art results. Other approaches for incorporating non-linearities considered replacing the linear interaction terms in the CPH model with deep neural networks, first explored in~\cite{faraggi1995neural}, followed by \cite{xiang2000comparison}, and again recently by~\cite{katzman2018deepsurv} with the \textit{DeepSurv} approach. Extensions to that work have involved convolutional neural networks and active learning for healthcare applications in oncology \cite{mobadersany2018predicting,nezhad2019deep}. However, those approaches are still subject to the same strong assumption of proportional hazards as the original CPH formulation.% Machine learning community has Instead of Deep Neural Networks, 

More recently, \cite{lee2018deephit,lee2019dynamic} proposed a deep learning approach, \textit{DeepHit}, to model the survival outcomes in the \textit{competing risks} scenario. Their approach is similar to ours in that they also aim to learn a fully parametric model, however their architecture only allows for the prediction of failure times over a discrete set of fixed size. This design may be impractical for problems with long survival horizons, in which precise prediction of actual failure times would require the discrete output space to be of high arity, resulting in a large number of parameters to be learnt, thus making parameter inference problematic in practice. Another practical limitation of this approach is that it can be more sensitive to events at shorter horizons and it may not model long term event horizons very well. In order to mitigate these issues, \cite{lee2019temporal} propose using black box optimization to adaptively select the best model from a large ensemble for a given event horizon. In this paper we explicitly demonstrate robust performance of our model at different quantiles of event times with varying extents of censoring.

Recent research also includes \textit{Deep Survival Analysis} proposed by \cite{ranganath16dsa}, which models survival problems with deep exponential families and aligns all observations by their failure time. 
Authors of \cite{chapfuwa2018adversarial} proposed using adversarial training methods by adapting a conditional Generative Adversarial Network \cite{mirza2014conditional} to survival regression problems. However, these approaches do not consider competing risks scenarios. 

In addition to the above mentioned approaches, non-parametric methods have also been popular for survival estimation. Published work includes improvements over the Kaplan-Meier (KM) estimator \cite{kaplan1958nonparametric} by fitting it in a small neighbourhood around an individual observation to accommodate conditioning. In addition, \cite{chen2019nearest} recently presented non-asymptotic error bounds with strong consistency results for these methods, and found that the use of forest ensembles for building conditional estimators of the survival function \cite{ishwaran2008random} can be an appropriate choice of kernel for such methods. Yet more recent approaches have involved Gaussian Processes \cite{alaa2017deep} with a similar intuition in the competing risks scenario.

\begin{figure*}[!htbp]
    \centering
    \includegraphics[width=0.75\textwidth, trim={0.5cm 6.15cm 0.6cm 1.6cm}, clip]{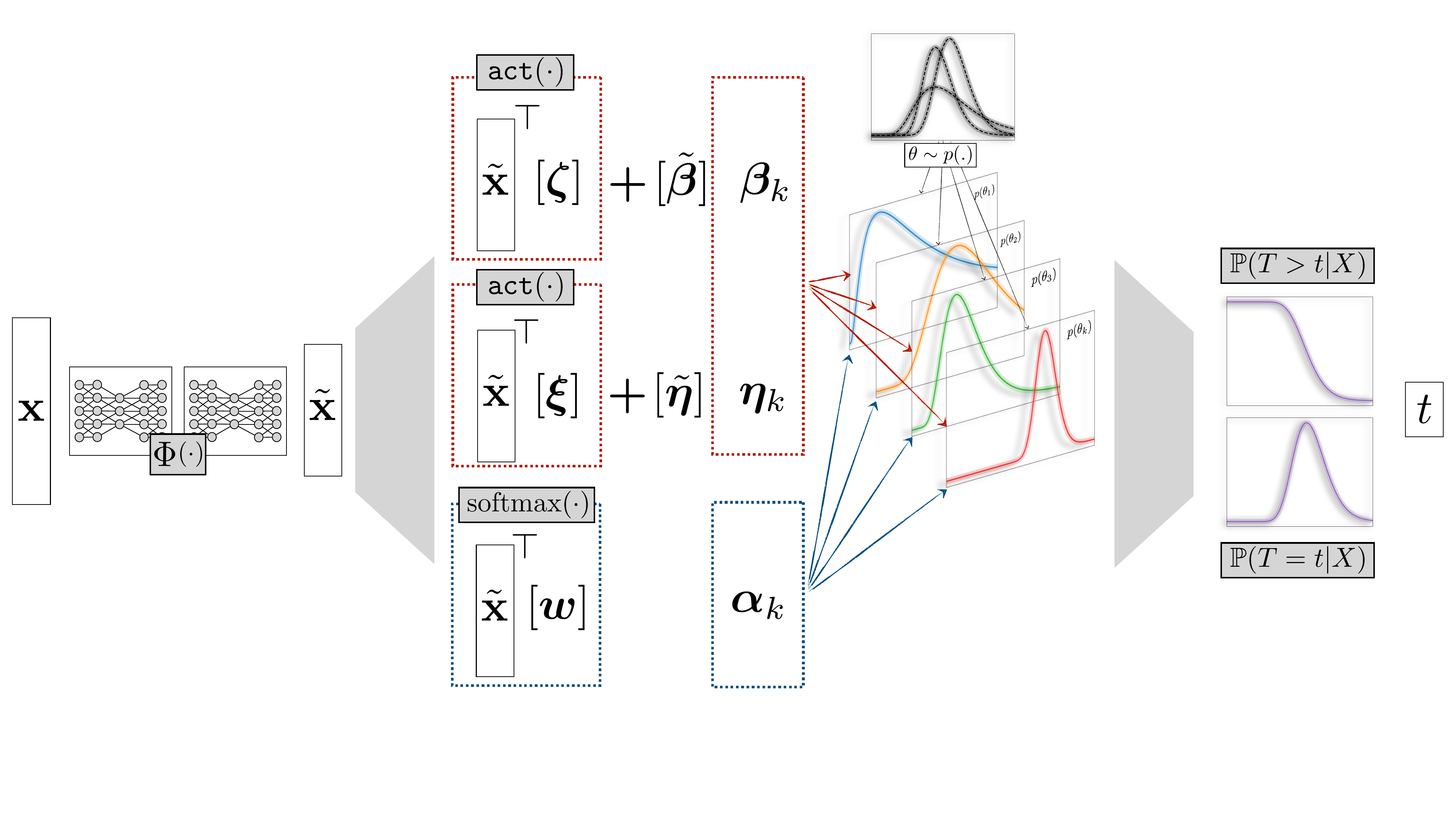}
    \caption{The proposed \method{} pipeline. The input features, $\mathbf{x}$ are passed through a deep multilayer perceptron followed by a softmax over mixture size, $K$. The Conditional Distribution of $\mathbb{P}(T|X=\mathbf{x})$ is then described as a mixture of $K$ \textsc{Primitive} distributions, drawn from some prior. }
    \label{fig:dsm}
\end{figure*}

Existing literature on survival regression can thus be grouped into two folds: 1) Semi-parametric approaches involving fitting proportional hazards (Coxian Models), and 2) Non-parametric models requiring some notion of similarity or kernel between individuals. 
To the best of our knowledge, the approach we propose here is the first fully-parametric method for survival regression in the presence of competing risks.

\section{ \large{Approach: `\method{}'}}

In this section we describe our approach, \method{} (\dsm{}) architecture and inference in further detail. Fig. \ref{fig:dsm} is a visual representation of our approach while Fig. \ref{sec:plate} describes the model in plate notation.

\subsection{Survival Data}

We assume that the survival data we have access to is \emph{right-censored}. This implies that our data, $\mathcal{D}$ is a set of tuples $\{(\mathbf{x}_i, t_i, \delta_i)\}_{i=1}^{N}$. Where typically, $\mathbf{x}_i \in \mathbb{R}^d$ are features associated with an individual i, $t_i$ is the time at which an event of interest took place, or the censoring time and $\delta_i$ is an indicator that signifies whether $t_i$ is event time or censoring time. For a given individual, we only either observe the actual failure or censoring time but not both. For simplicity, it is assumed that the true data generating process is such that the censoring process is independent of the actual time to failure. We denote the uncensored subset $(\delta=1)$ of data as $\mathcal{D}_U$ and the censored $(\delta=0)$ subset as $\mathcal{D}_C$.

\subsection{Primitive Distributions}

We choose to model the conditional distribution $\mathbb{P}(T|X=\mathbf{x})$ as a mixture over $K$ well-defined, parametric distributions which we call as \textsc{Primitive} distributions for the remainder of this paper. Given that we are modelling survival times, a natural assumption for these \textsc{Primitive} distributions is to have support only in the space of positive reals. Another property of interest is to have a closed form solution for the \textsc{cdf}, as this would enable the use of gradient based optimization for Maximum Likelihood Estimation. 

\begin{table}[!htbp]
\centering
\caption{Choices for the \textsc{Primitive} distributions.}

\label{tab:dists}
\vspace{0.1in}
\begin{tabular}{c|c|c}
\toprule  \midrule
     & \textsc{Weibull} & \textsc{Log-Normal} \\ \midrule       
     $\textsc{pdf}(t)$ & $\frac{\eta}{\beta} \big( \frac{t}{\beta}\big) ^{\eta-1}  e^{ - \big( \frac{t}{\beta} \big)^{\eta} }$ & $\frac{1}{t\beta\sqrt{2\pi}}e^{ - \frac{(\ln{t} - \eta)^2}{2\beta^2}}$ \\
     $\textsc{cdf}(t)$ &  $e^{  - \big( \frac{t}{\beta} \big)^{\eta}}$ & $ \frac{1}{2}\text{erfc}{}\Big(-\frac{\ln{t} - \eta}{\sqrt{2}\beta}\Big)$ \\
     % \midrule
     % & \textsc{Log-Normal} \\ \midrule
     % $f_\mathcal{T}(t)$ & $\frac{1}{t\beta\sqrt{2\pi}}e^{ - \frac{(\ln{t} - \eta)^2}{2\beta^2}}$ \\
     % $S_\mathcal{T}(t)$ &  $\frac{1}{2} - \frac{1}{2}\erf{}\big(\frac{\ln{t} - \eta}{\sqrt{2}\beta}\big)$\\
     \midrule \bottomrule
\end{tabular}
\vspace{-0.1in}
\end{table}

 For  \dsm{}, we experiment with two types of distributions that satisfy this property, the Weibull and the Log-Normal distribution. The first of them has closed form \textsc{pdf} and \textsc{cdf}. For the Log-Normal, we compute the \textsc{cdf} by using the standard approximation of the complementary error function \texttt{erfc} as implemented in \texttt{PyTorch}. The full functional forms of the distributions are listed in Table \ref{tab:dists}. We parameterize the $\beta_k$ and $\eta_k$  as:
$$\beta_k  = \Tilde{\beta}_k + \texttt{act}(\Phi_\theta(\bm{x}_i)^{\top}\bm{\zeta} ),$$
$$\eta_k = \Tilde{\eta_k} + \texttt{act}(\Phi_\theta(\bm{x}_i)^{\top}\bm{\xi})$$

Here the \texttt{act}$(\cdot)$ is the \textsf{SELU} and \textsf{Tanh} activation functions for the Weibull and Log-Normal respectively, and $\mathbf{\Phi}(.)$ is a Multilayer Perceptron (MLP). 
$\mathbf{x}_i$ are the input covariates. $\{ \theta, \bm{\xi},\bm{\zeta}, \beta$ and $\eta \}$ are all parameters that are learnt during training. Another set of parameters that are learnt are $\bm{w}$ that determine the mixture weights for each data point. 

Figure~\ref{sec:plate} introduces the proposed model in plate notation and the corresponding conditional independence assumptions of the Graphical Model.
The input features, $\bm{x}_i$, are passed through the MLP $\mathbf{\Phi_\theta}$ to determine the representation $\widetilde{\bm{x}}_i$. This representation then interacts with the additional set of parameters to determine the mixture weights $\bm{w}$ and the parameters of each of $K$ underlying survival distributions $\{ \eta_k, \beta_k \}_{k=1}^{K}$. The final individual survival distribution for the event time $T$ is a weighted average over these $K$ distributions.
\begin{tcolorbox}[enhanced, sharp corners, boxrule=.8pt,drop fuzzy shadow,colback=white,colframe=black, ,title={}]
    \centerline{  
    \includegraphics[width=0.85\linewidth]{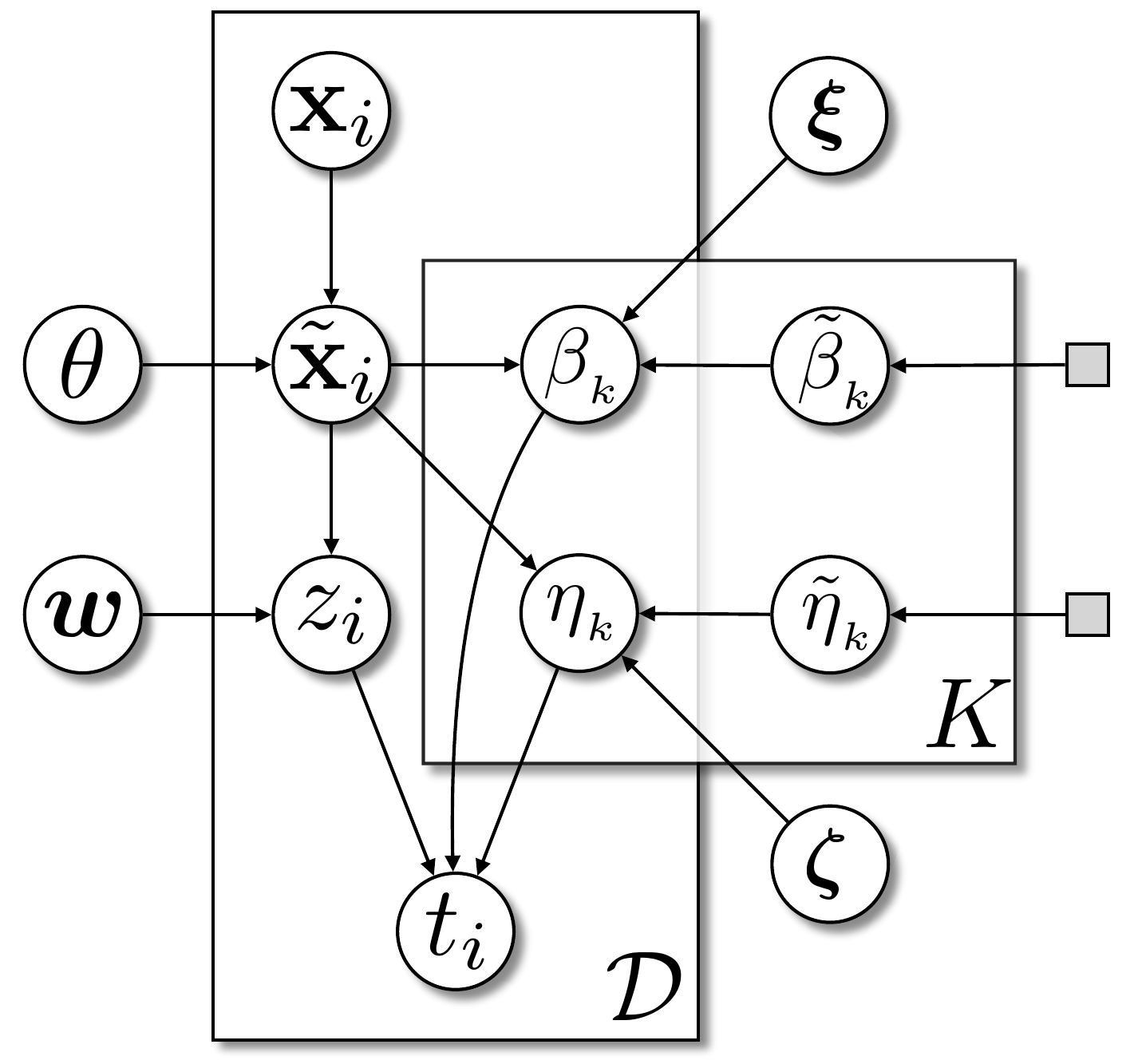}
    \label{fig:plate}
}   
    \vspace{1em}
    \textbf{The Generative Story}
    \begin{enumerate}
    \item $\mathbf{x}_i \sim \mathcal{D}$\\
     We draw the co-variates of the individual, $\mathbf{x}_i$
     \item $\bm{w}, \bm{\zeta}, \bm{\xi} \sim \mathcal{N}(0, \nicefrac{1}{\lambda})$\\
     The parameters of the model are drawn from a zero mean Gaussian distribution.
    % \item $z_i \sim \text{Discrete}(\textsc{softmax}(\textsc{MLP} (\mathbf{x}_i, \mathbf{w}))) $\\ Conditioned on the covariates, $\mathbf{x}_i$ and the parameters, $\mathbf{w} $ we draw the latent $z_i$
    \item $z_i \sim \text{Discrete}\big(\texttt{softmax}({\Phi_\theta}(\bm{x}_i)^{\top}\bm{w} )\big) $\\ Conditioned on the covariates, $\bm{x}_i$ and the parameters, $\bm{w} $ we draw the latent $z_i$
    \item $ \log \Tilde{\beta}_k \sim \mathcal{N}( \beta_0, \nicefrac{1}{\lambda})$\\ $\log \Tilde{\eta}_k \sim \mathcal{N}(\eta_0, \nicefrac{1}{\lambda}) $\\ The set of parameters $\{\Tilde{\beta}_k\}_{k=1}^{K}$ and $\{\Tilde{\eta}_k\}_{k=1}^{K}$ are drawn from the prior $\beta_0$ and $\eta_0$.
    % \item $\begin{aligned}\bm{t}_i \sim \textsc{Primitive}\big(\beta_k &+ \texttt{act}(\Phi_\theta(\bm{x}_i)^{\top}\bm{\zeta} ),\\
    %  \eta_k &+ \texttt{act}(\Phi_\theta(\bm{x}_i)^{\top}\bm{\xi} )\big)\end{aligned}$\\ Finally, the event time $t_i$ is drawn conditioned on $\beta_{z_i}$ and $\eta_{z_i}$
     \item $\begin{aligned}\bm{t}_i \sim \textsc{Primitive}&\big(\beta_k,\eta_k)\\ \text{where, } \beta_k  &= \Tilde{\beta}_k + \texttt{act}(\Phi_\theta(\bm{x}_i)^{\top}\bm{\zeta} )\\ \eta_k &= \Tilde{\eta_k} + \texttt{act}(\Phi_\theta(\bm{x}_i)^{\top}\bm{\xi} )\end{aligned}$\\ Finally, the event time $t_i$ is drawn conditioned on $\beta_{z_i}$ and $\eta_{z_i}$.
    \end{enumerate}

\captionof{figure}{\method{} in plate notation.}
  \label{sec:plate}
\end{tcolorbox}

%\newpage

\newpage

\subsection{Parameter Estimation}
In order to accommodate for heterogeneity arising in the data, we propose to model the survival distribution of each individual as a fixed size mixture of survival distribution primitives. At test time, the survival function corresponding to this held out individual is described as a weighted mixture of the survival distribution primitives. Here, the weights are a softmax of the output of a deep neural network. At training time, the parameters of the network and the survival distribution primitives are learnt jointly.

\noindent \textbf{Uncensored Loss.}
We consider the maximum likelihood estimator for the uncensored data which can be written as:
\begin{align}
    \nonumber \ln \mathbb{P}(&\mathcal{D}_{U}|  \mathbf{\Theta}) = \ln \bigg(\prod_{i=1}^{\mathcal{|D|}} \mathbb{P} (T=t_i|X=\mathbf{x}_i, \mathbf{\Theta})\bigg)\\
    \nonumber&=\sum_{i=1}^{\mathcal{|D|}}\ln \bigg( \sum_{k=1}^{K} \mathbb{P} (T=t_i|Z, \beta_k, \eta_k)\mathbb{P} (Z | X=\mathbf{x}_i, \bm{w}) \bigg)\\
    \nonumber&=\sum_{i=1}^{|\mathcal{D}|}\ln \bigg( \mathop{\mathbb{E}}\limits_{Z\sim (\cdot| \mathbf{x}_i, \bm{w} )} [\mathbb{P} (T=t_i|Z, \beta_k, \eta_k)] \bigg)\\
    \nonumber&(\text{Applying Jensen's Inequality})\\
    \nonumber&\geq  \sum_{i=1}^{|\mathcal{D}|} \bigg( \mathop{\mathbb{E}}\limits_{Z\sim (\cdot| \mathbf{x}_i, \bm{w} )} [ \ln \mathbb{P} (T=t_i|Z, \beta_k, \eta_k)] \bigg)\\
    \nonumber&\triangleq \textbf{ELBO}_{U}(\Theta)
\end{align}

\noindent \textbf{Censoring Loss.}
Proceeding as above, we can write the lower bound of loss for the censored observations as:
\begin{align}
\nonumber\ln \mathbb{P}(\mathcal{D}_{C}| \Theta) &= \ln \bigg(\prod_{i=1}^{\mathcal{|D|}} \mathbb{P} (T>t_i|X=\mathbf{x}_i, {\Theta})\bigg) \\
 \nonumber&\geq  \sum_{i=1}^{|\mathcal{D}|} \bigg( \mathop{\mathbb{E}}\limits_{Z\sim (\cdot| \mathbf{x}_i, w )} [ \ln \mathbb{P} (T> t_i|Z, \beta_k, \eta_k)] \bigg)\\
    \nonumber&\triangleq  \textbf{ELBO}_{C}(\Theta)
\end{align}
% However we also have censored individuals to be accounted for in the MLE. In the presence of the censored individuals, the MLE can be rewritten as 
% \begin{align}
%     L_{\text{U}} &= \prod_{i=1}^{\mathcal{|D|}} P (T>t_i|X=\mathbf{x}_i, {\Theta})\\
%     &=\prod_{i=1}^{\mathcal{|D|}}\sum_{k=1}^{K} P (T>t_i|Z, \beta_k, \eta_k)P (Z | X=\mathbf{x}_i, w)\\
%     &=\sum_{\mathcal{D}} \text{LL}(\Theta; x)
% \end{align}

\noindent \textbf{Mitigating Long Tail Bias.}
Survival distributions with positive support typically have long tails, a complication that adds to the bias when performing Maximum Likelihood Estimation. Note that for the censored instances of data, we are maximizing the probability $\mathbb{P}(T>t)$. One conceivable way of adjusting for the long-tail bias is to instead maximize $\mathbb{P}(t_{\text{max}}>T>t)=\mathbb{P}(T>t)-\mathbb{P}(T>t_{\text{max}}) $ where $t_\text{max}$ is some arbitrarily large value that can be tuned as a hyper-parameter. However, for simplicity, we choose to directly discount the censoring loss by multiplying it with a factor $\alpha \in [0, 1]$, which has a similar effect of diminishing bias arising from the long tails.

\noindent \textbf{Prior Loss.}
We include the strength of the prior on the $\beta_k$, $\eta_k$ as:
\begin{align}
    \nonumber\mathcal{L}_{\text{prior}} &= \ln \bigg( \prod_{k=1}^{K} \mathbb{P} (\beta_k, \eta_k| \beta, \eta) \bigg)\\
    \nonumber&=\sum_{k=1}^{K} \ln \mathbb{P} (\beta_k| \beta) + \ln \mathbb{P} ( \eta_k| \eta) \\
    \nonumber&=\lambda \sum_{k=1}^{K} || \beta_k -\beta ||_{_{\mathbf{2}}}^{2} + || \eta_k -\eta ||_{_{\mathbf{2}}}^{2}
\end{align}

\noindent \textbf{Combined Loss.}
We finally combine the individual components of loss described above into:
$$ \mathcal{L}_{\text{combined}} =  \textbf{ELBO}_{U}(\Theta) + \alpha \cdot \textbf{ELBO}_{C}(\Theta)  + \mathcal{L}_{\text{prior}}   $$

Here, $\alpha$ is a scalar hyperparameter that trades off the contribution of regression loss vis-à-vis the evidence lower bound of the uncensored observations to the combined objective function. For a complete formulation of the loss function, in terms of functions and parameters please refer to Appendix \ref{apx:loss}.

% (\%age)
\begin{table*}[!ht]
    \centering
       \caption{Descriptive statistics of the datasets used in the experiments.}
           \label{tab:my_label}
\vspace{0.1in}
    \begin{tabular}{l|l|r|r|r|r|r}
        \toprule \midrule
        Dataset& Type & Dataset Dim. & Feature Dim. & \multicolumn{2}{c|}{No. Events} & No. Censoring  \\ \midrule
         \textbf{SUPPORT} & Single Risk& 9,105 & 30 & \multicolumn{2}{c|}{6,201 (68.1 \%)} & 2,904 (31.9 \%) \\ \midrule
         \textbf{METABRIC} & Single Risk & 1,904 & 9 & \multicolumn{2}{c|}{1,103 (57.9 \%)} & 801 (42.1 \%) \\ \midrule
         % \textbf{MAGGIC}&Single Risk& \\ \hline
         \textbf{SYNTHETIC} & Competing Risks & 30,000 & 12 & \makecell[l]{Event 1 \\Event 2} & \makecell[r]{7,600 (25.3\%) \\  7,400 (24.7\%)} & 15,000 (50.0 \%)\\ \midrule
         \textbf{SEER} & Competing Risks & 65,481 & 21 & \makecell[l]{BC \\CVD} & \makecell[r]{13,564 (20.7\%) \\4,245 \hspace{.15em} (6.5\%)} & 47,672 (72.8 \%)\\\midrule
         \bottomrule
         
    \end{tabular}
 \vspace{-0.1in}
\end{table*}

\subsection{Handling Multiple \textit{Competing Risks}}

We adapt \method{} to scenarios involving multiple competing risks by allowing learning of a common representation for the multiple risks by passing through a single MLP ($\bm{\Phi}(.)$ in Fig. \ref{fig:dsm}). This representation then interacts with a separate set of $\{ \bm{\xi}, \bm{\zeta}, \bm{w}  \}$ in order to describe the event distribution for each competing risk. Maximum Likelihood Estimation is performed by treating the occurrence of a competing event before the other event as a form of independent censoring. This strategy allows the model to leverage knowledge from the two (in general, more than two) competing tasks by allowing parameter sharing through a single intermediate representation.

%\todo[inline]{FILL IN SOMETHING HERE!}

\section{Experiments}

\label{sec:experiments}

We evaluate \method{} on their ability to measure relative risks for a single event of interest in the presence of censoring, and then we further consider ablation experiments where we artificially increase the amount of censoring to demonstrate the robustness of the proposed approach. Finally, we demonstrate \dsm{}'s ability to learn representations of the covariates for transferring knowledge across two events in the \textit{competing risks} scenario with censoring.

\subsection{Datasets}

\noindent \textbf{Single Event/Single Risk.}
We evaluated performance of the proposed method on the following real-world medical datasets with single events: Study to Understand Prognoses Preferences Outcomes
and Risks of Treatment (SUPPORT) \cite{knaus1995support}, and Molecular Taxonomy of Breast Cancer International Consortium (METABRIC) \cite{curtis_2012}. A brief introduction of each dataset is provided below, and the details of preprocessing the data are provided in Appendix \ref{apx:preprocessing}.

\textbf{SUPPORT:} The SUPPORT resulted from a study conducted to describe a prognostic model to estimate survival over a 180-day period for 9,105 seriously ill hospitalized patients. Of the 9,105 patients, 6,201 patients (68.1\%) were followed to death, with a median survival time of 58 days. We used 30 patient covariates, including age, gender, race, education, income, physiological measurements, co-morbidity information, etc. Missing values of certain physiological measurements were imputed using the suggested normal values\footnote{ \url{http://biostat.mc.vanderbilt.edu/wiki/Main/SupportDesc}} and other missing values were imputed using the mean value for numerical features and the mode for categorical features. 

\textbf{METABRIC:} The METABRIC came from a study conducted to determine new breast cancer subgroups and facilitate treatment improvement using patients' gene expressions and clinical variables. The dataset consists of 1,904 patients and 9 features. 1,103 patients (57.9\%) were followed to death with a median survival time of 115.9 months. The dataset used was preprocessed as in \cite{katzman2018deepsurv} and downloaded from the \texttt{PySurvival} library\footnote{\url{https://square.github.io/pysurvival/}}.

\noindent \textbf{Competing Risks.}
We also evaluated the performances on two datasets with competing risks: a synthetic dataset and the Surveillance, Epidemiology, and End Results (SEER) dataset.

\textbf{SYNTHETIC:} In order to demonstrate the effectiveness of \dsm{} as a representation learning framework, we experiment with synthetic data that is generated following the spirit of \cite{alaa2017deep} \&
\cite{lee2018deephit} using the same generative process as they described. 
\begin{align*}
    \mathbf{x}^{(i)}_1,\mathbf{x}^{(i)}_2, \mathbf{x}^{(i)}_3 &\sim \mathcal{N}(0, \mathbf{I})\\
    T_1^{(i)} &\sim \exp\bigg( (\gamma_3^\top \mathbf{x}^{(i)}_3)^2 + \gamma_1^\top \mathbf{x}^{(i)}_1 \bigg)\\
    T_2^{(i)} &\sim \exp\bigg( (\gamma_3^\top \mathbf{x}^{(i)}_3)^2 + \gamma_2^\top \mathbf{x}^{(i)}_2 \bigg)
\end{align*}

Here $\mathbf{x}^{(i)} =  ( \mathbf{x}_1^{(i)}, \mathbf{x}_2^{(i)} , \mathbf{x}_3^{(i)})$ is a tuple representing the covariates of the individual $i$. The Event times $T_1$ and $T_2$ are exponentially distributed around functions that are both linear and quadratic in $X$. We generate 30,000 patients from the distribution out of which 50\% are subjected to random right censoring by uniformly sampling the censoring times in the interval $[0, \text{min}\{T_1, T_2 \}]$. Clearly, the choice of our distributions for the event times are not independent, and would allow a model to leverage knowledge of one event to better predict the other, which is what we intend to demonstrate. 

\textbf{SEER:} This dataset\footnote{\url{https://seer.cancer.gov/}} provides information on cancer statistics among the United States population. We focused on the breast cancer patients in the registries of Alaska, San Jose-Monterey, Los Angeles and rural Georgia during the years from 1992 to 2007, with the follow-up period restricted to 10 years. Among the 65,481 patients, 13,564 (20.7\%) died due to breast cancer (BC) and 4,245 (6.5\%) died due to cardiovascular disease (CVD), which were treated as the two competing risks in our experiments. We used 21 patient covariates, including age, race, gender, diagnostic confirmation, morphology information (primary site, laterality, histologic type, etc.), tumor information (size, type, number etc.), and surgery information. Missing values were imputed using the mean value for numerical features and the mode for categorical features. 

\subsection{Baselines}

We compare the performance of \dsm{} to the following competing baseline approaches:

\textbf{Cox Proportional Hazards (CPH):} This is the standard semi-parametric model, making the assumption of constant baseline hazard. The features interact with the learnt set of weights in a log-linear fashion in order to determine the hazard for a held out individual. 

\textbf{Random Survival Forests (RSF):} This is a popular non-parametric approach involving learning an ensemble of trees, adapted to censored survival data~\cite{ishwaran2008random}.

\textbf{DeepSurv (DS):} Proposed by \cite{katzman2018deepsurv}, DeepSurv involves learning a non-linear function that describes the relative hazard of a test instance. It makes the familiar assumption of constant baseline hazard, as does \textbf{CPH}.

\textbf{DeepHit (DH)} ~\cite{lee2018deephit}: This approach involves learning the joint distribution of all event times by jointly modelling all competing risks and discretizing the output space of event times. 

\textbf{Fine-Gray (FG)} ~\cite{fine1999proportional}: This is a classic approach used for modelling competing risks that focuses on the cumulative incidence function by extending the proportional hazards model to sub-distributions.

For the SYNTHETIC and SEER datasets with competing risks, we compare performance of \dsm{} to cause-specific (\textbf{cs-}) versions of \text{CPH} and \text{RSF} that involve learning separate survival regressions for each competing event by treating the other event as censored. 

\subsection{Performance Metrics}

We evaluate \dsm{} by assessing the ordering of pairwise relative risks using Concordance-Index (C-Index) \cite{harrell_1982}. To demonstrate performance of our approach vs.\ the methods subject to Coxian assumption, we show the comparison of performances using the time-dependent Concordance-Index $C^{td}$ \cite{antolini2005time}.
\begin{align*}
C^{td }(t) = \mathbb{P}\big( \hat{F}(t| \mathbf{x}_i) > \hat{F}(t| \mathbf{x}_j)  | \delta_i=1, T_i<T_j, T_i \leq t \big) 
\end{align*}
Here, $\hat{F}(t| X)$ is the estimated \textsc{CDF} by the model at the truncation time $t$, given features $X$. The probability is estimated by comparing relative risks pair-wise. In order to obtain an unbiased estimate for the quantity, we adjust the estimate with an inverse propensity of censoring estimate, as is common practice in survival analysis literature \cite{gerds2013estimating}.

Observing $C^{td}$ by different evaluation time horizons enable us to measure how good the models are at capturing the possible changes in risk over time, thus alleviating the restrictive assumption C-Index makes of constant proportional hazards. For completeness, we report the $C^{td}$ at different truncation event horizon quantiles of 25\%, 50\%, 75\%.

Practical utility of deployed applications and their interpretability requires survival analysis models to be well calibrated. 
We thus further assess calibration of DSM in comparison to the baselines by computing the Censoring Weighted Brier Score~\cite{gerds2006consistent, graf1999assessment} at each event's quantile.

\subsection{Experimental Setup}
\label{sec:experiment}

\noindent \textbf{Hyperparameters:} For all the experiments described subsequently we train \dsm{} with the \textsf{Adam} optimizer \cite{kingma2014adam} using learning rates of $\{1\times10^{-3}, 1\times10^{-4}\} $. The number of experts, $K$ for each event is chosen from $\{ 4, 6, 8 \}$ and the discounting factor $\alpha$ is chosen from $\{\nicefrac{1}{2}, \nicefrac{3}{4},1\}$. The prior strength $\lambda$ is set as $1\times10^{-8}$ for all the experiments and not tuned. We report the $C^{td}$ for the best performing set of parameters over the grid in cross validation for both DSM and the baselines. The representation learning function $\mathbf{\Phi}(.)$ is a fully connected Multi-Layer Perceptron with 1 or 2 Hidden Layers with the number of nodes in $\{ 50, 100 \}$ and ReLU6 activations. The choice of Log-Normal or Weibull outcome is further tuned as a hyper parameter. All experiments were conducted in \texttt{PyTorch}\cite{paszke2019pytorch}.

\noindent \textbf{Evaluation Protocol:} For each experiment we report the standard error around the mean of the $C^{\text{td}}$ in 5-fold cross validation.\footnote{Except for METABRIC, where we perform 10-fold cross validation to get tighter confidence bounds.} For full details of hyperparameter choices for the baselines please refer to the Appendix \ref{apx:baselines}.

\begin{figure*}[!htbp]
    \centering
    \begin{minipage}{0.33\textwidth}
        \includegraphics[width=\textwidth]{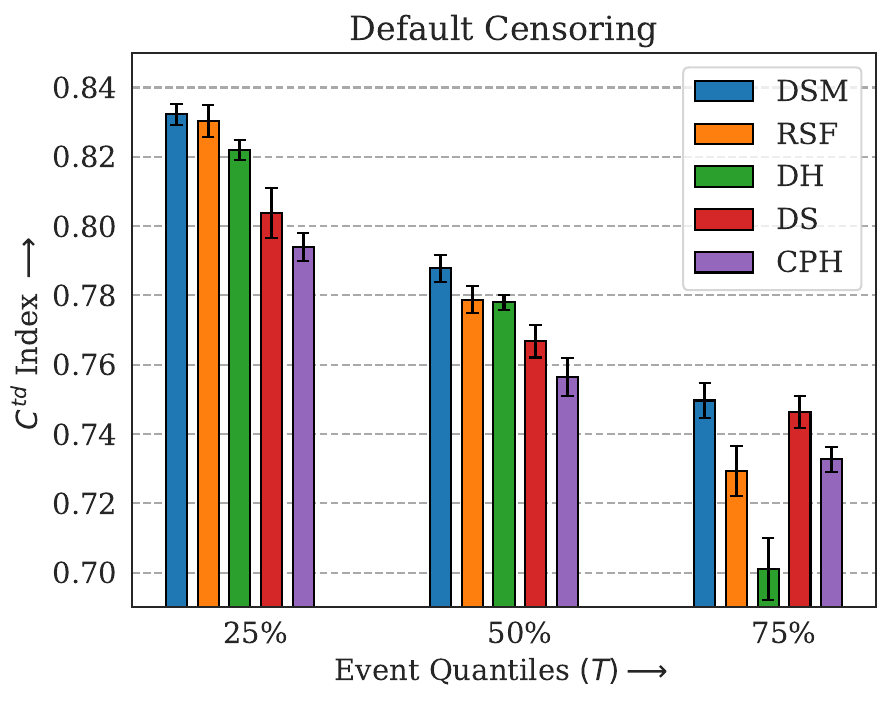}
    \end{minipage}%
    \begin{minipage}{0.33\textwidth}
        \includegraphics[width=\textwidth]{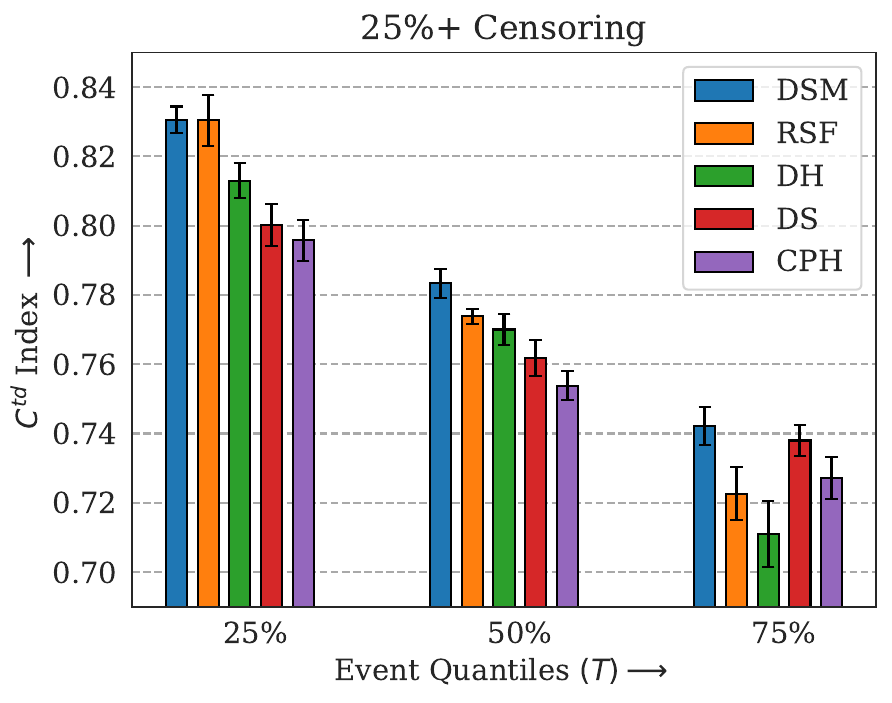}
    \end{minipage}%
        \begin{minipage}{0.33\textwidth}
        \includegraphics[width=\textwidth]{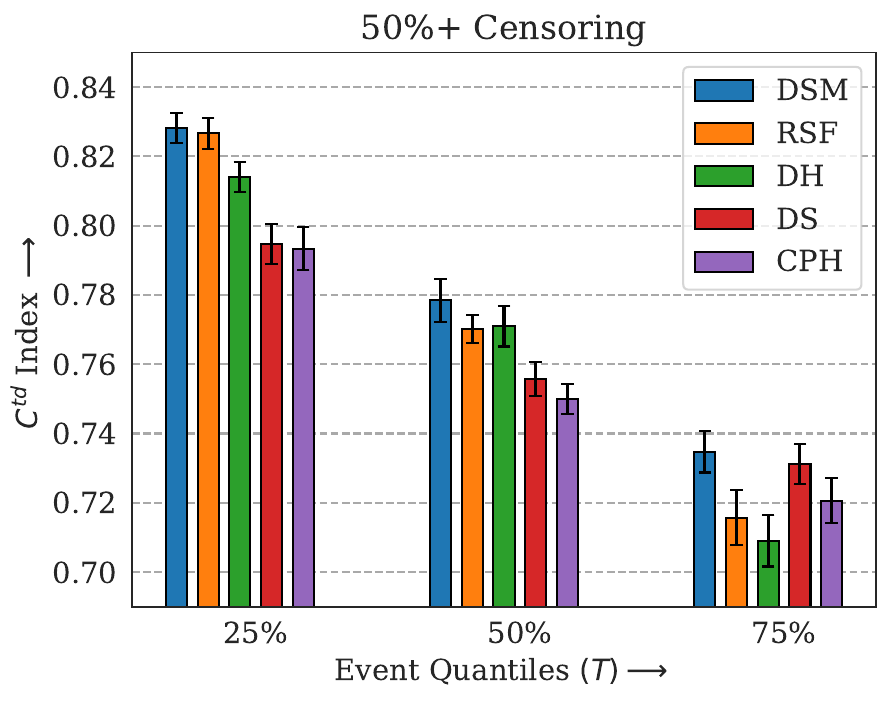}
    \end{minipage}%
    % \begin{minipage}{0.25\textwidth}
    %     \includegraphics[width=\textwidth]{figs/support/support_75.pdf}
    % \end{minipage}
    \caption{$C^{td}$ for \textbf{SUPPORT} dataset at different quantiles of event times for different levels of censoring.}
    \label{fig:support}
\end{figure*}

\begin{figure*}[!t]
    \centering
    \begin{minipage}{0.33\textwidth}
        \includegraphics[width=\textwidth]{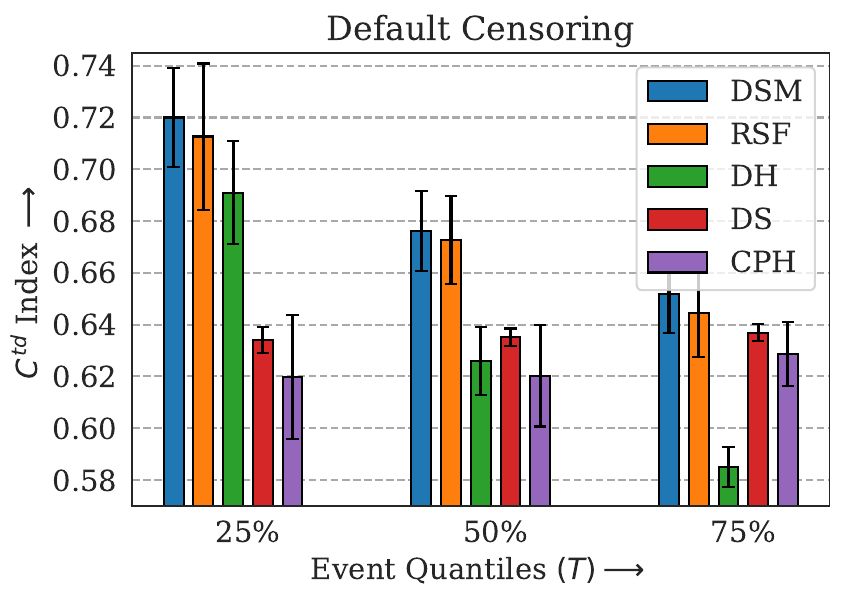}
    \end{minipage}%
    \begin{minipage}{0.33\textwidth}
        \includegraphics[width=\textwidth]{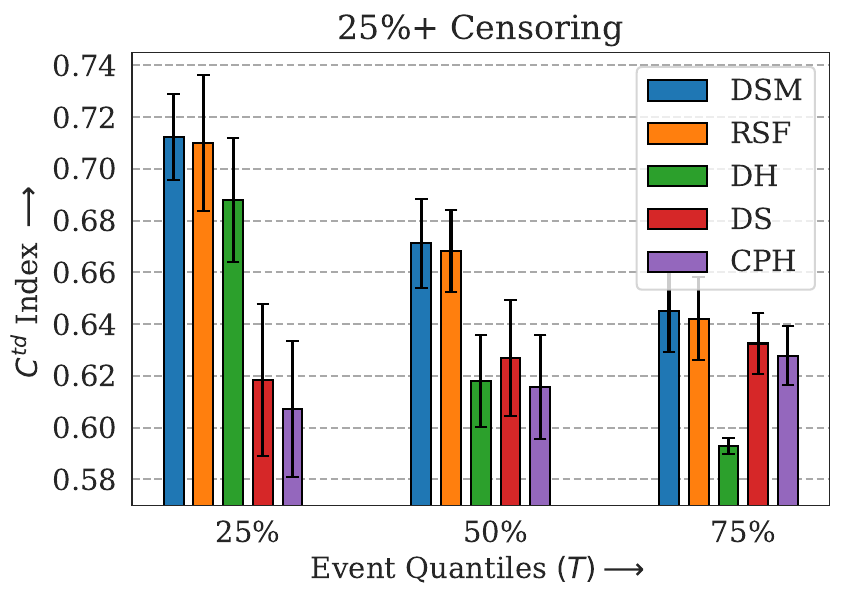}
    \end{minipage}%
    \begin{minipage}{0.33\textwidth}
        \includegraphics[width=\textwidth]{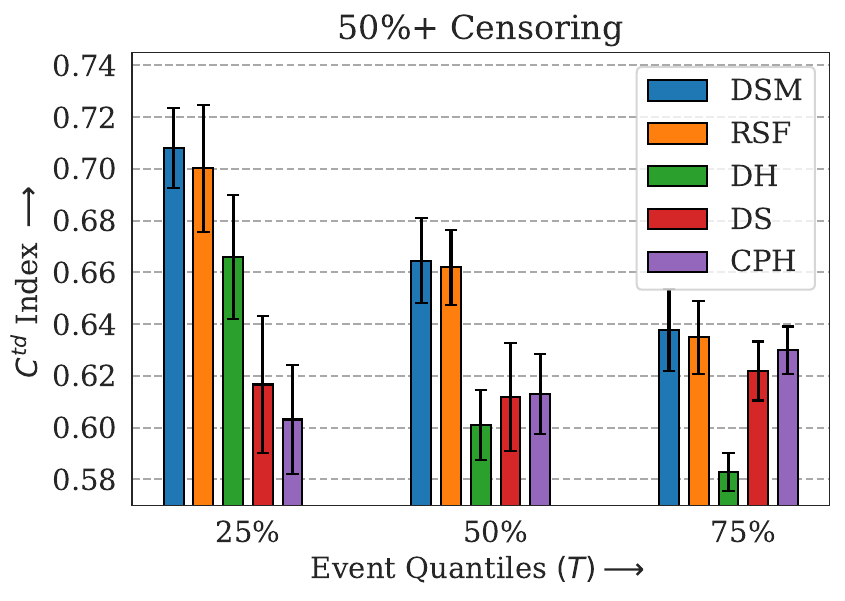}
    \end{minipage}%

    \caption{$C^{td}$ for \textbf{METABRIC} dataset at different quantiles of event times for different levels of censoring.}
    \label{fig:metabric}
\end{figure*}

\begin{figure*}[!t]
\begin{minipage}[!t]{0.5\textwidth}
    \centering
    \begin{minipage}{0.5\textwidth}
    \includegraphics[width=\textwidth]{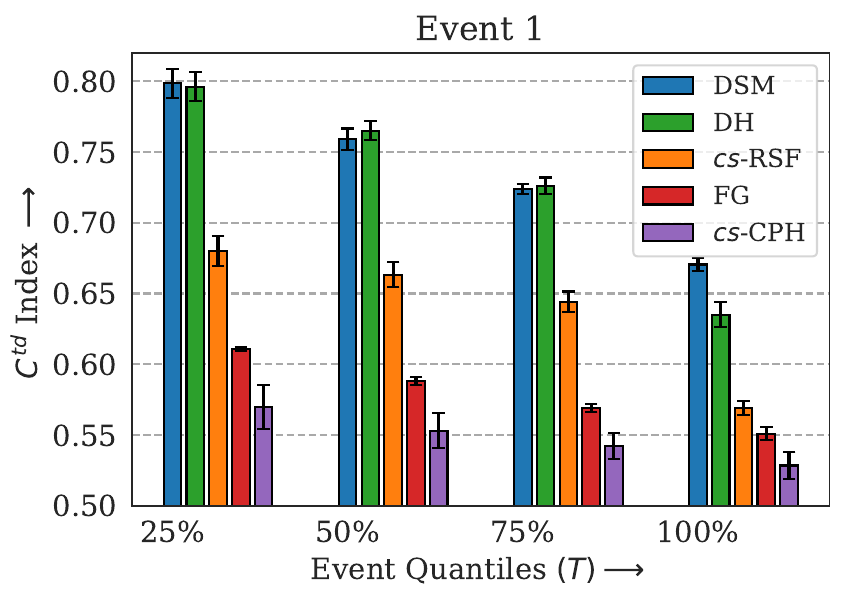}
    \end{minipage}%
    \begin{minipage}{0.5\textwidth}
    \includegraphics[width=\textwidth]{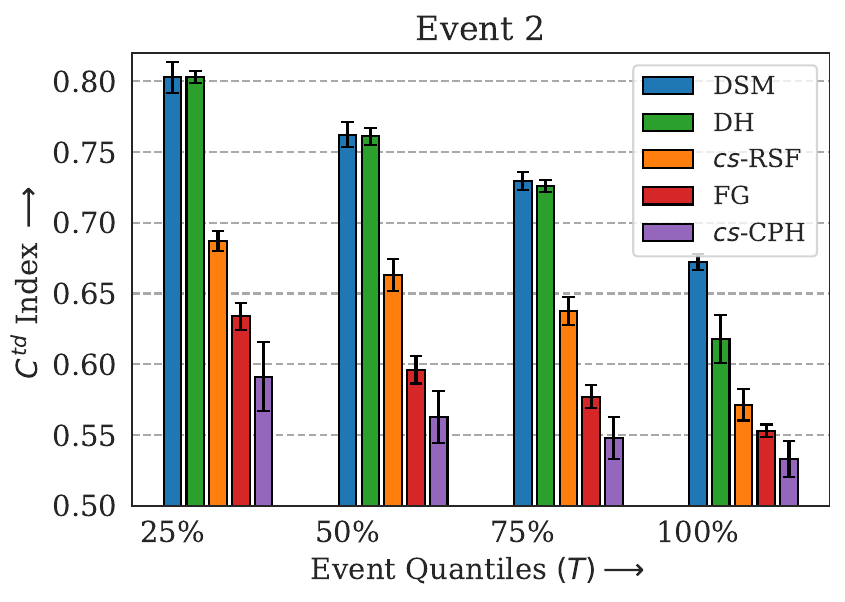}
    \end{minipage}
    \captionof{figure}{$C^{td}$ for competing risks on \textbf{SYNTHETIC}}
    \label{fig:synthetic}
\end{minipage}%
\begin{minipage}[!t]{0.5\textwidth}
    \centering
    \begin{minipage}{0.5\textwidth}
    \includegraphics[width=\textwidth]{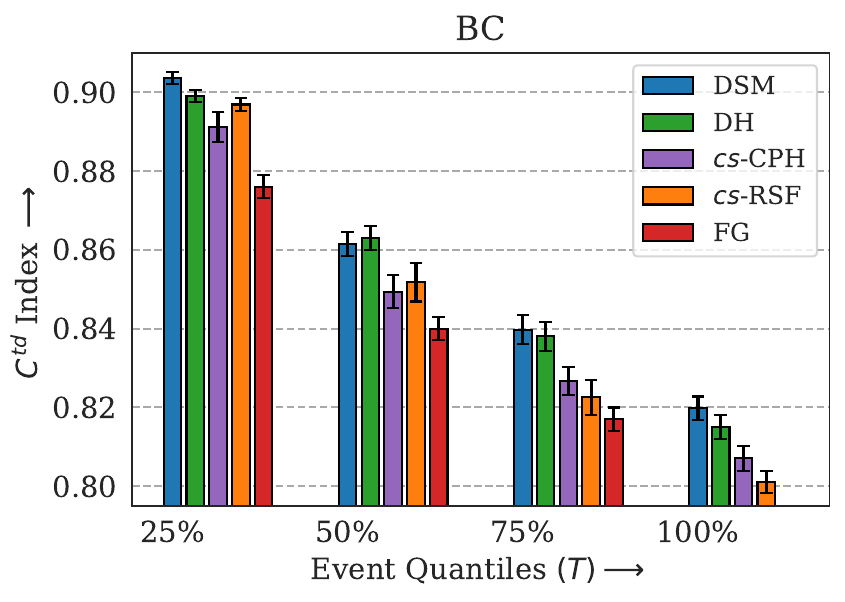}
    \end{minipage}%
    \begin{minipage}{0.5\textwidth}
    \includegraphics[width=\textwidth]{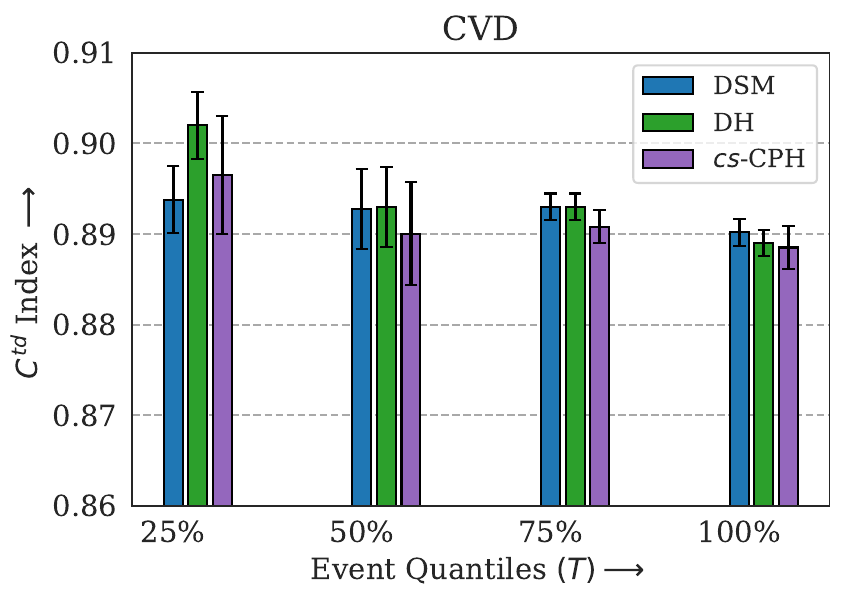}
    \end{minipage}
    \captionof{figure}{$C^{td}$ for competing risks on \textbf{SEER}}
    \label{fig:SEER}
\end{minipage}
\end{figure*}

\subsection{Single Event Survival Regression}
\label{sec:single}

%\subsection{Robustness to Censoring}
Parameter inference for \dsm{} involves the exploitation of a closed form of the \textsc{CDF}, which makes \dsm{} amenable to gradient based optimization. Naturally one would expect that a greater amount of censoring will reduce the available information to be modeled, thus adding bias and leading to poorer estimates of the survival function. % thus adversely affecting the estimation problem. 

In this section we will empirically investigate \dsm{}'s robustness to censoring and compare it to the relevant baselines by artificially censoring the event times. We uniformly sample a censoring time between $[0, T)$ for a randomly chosen subset of the uncensored training data. This is only applied to the uncensored instances of the training splits with the same experimental protocol as used in the previous Section \ref{sec:experiment}. (By not censoring the test splits we are able to better estimate the $C^{td}$). We perform this artificial censoring on the single event \textsc{METABRIC} and \textsc{SUPPORT} datasets and reduce the uncensored training data to 50\% and 25\% of its original amount. 

%\subsection{Discussion}
Figure \ref{fig:support} summarizes the performance of \dsm{} on the SUPPORT dataset in 5-fold cross validation. Notice that RSF is comparable to DSM in the 25\% quantile of event time horizons across all levels of censoring, however DSM significantly outperforms RSF on the longer event quantiles. Similarly we observed that although DeepSurv was competitive in longer event horizons, \dsm{} significantly outperformed DeepSurv in the shorter horizons, demonstrating superiority.

For METABRIC, we observed that DSM outperformed the Deep Learning baselines significantly. Although RSF was competitive, DSM outperformed RSF on average in 10-fold cross validation. 

{Brier Scores and $C^{\text{td}}$ obtained for METABRIC and SUPPORT data, can be found in Appendix~\ref{apx:results}, and they present \dsm{}'s calibration ability to be at least on-par and often better when compared to the alternatives.}

We close this section with a brief discussion on the scenarios when DSM can possibly achieve performance gain over alternative methods. When the data is sparse during certain time spans, e.g. the longer event horizons, the parametric nature of DSM makes it more robust than the non-parametric or deep learning based methods. On the contrary, the fitting of other competing non-parametric models and deep learning based approaches suffer from this lack of data. Furthermore, since DSM does not make the constant proportional hazard assumption like the CPH model or its variants, DSM is able to capture the flexible patterns of the survival functions when such assumptions do not hold.

\subsection{Competing Risks Scenario}

For the SYNTHETIC dataset, we observe in Fig.~\ref{fig:synthetic} that \dsm{} is competitive with DeepHit and outperforms all the other baselines in the 25\%, 50\%, 75\% quantiles of event horizons. For comparison, we also report the performance at 100\% quantile and observe that \dsm{} is significantly superior to DeepHit for both events, thus confirming its robustness to events at longer horizons.

From Fig.~\ref{fig:SEER}, on the SEER dataset we observe that for the majority risk, Breast Cancer, DSM significantly outperformed all the other baselines. The results for CVD were less conclusive with DeepHit being competitive at the 25\% quantile. We owe this to the class imbalance between the two types of risks. Note that for visual clarity we do not report Fine-Gray and cs-RSF since their performance was poor. We defer the actual numbers and confidence intervals to Appendix~\ref{apx:results}.

\section{Representation Learning and Knowledge Transfer}

We also conducted a set of experiments to evaluate the performance of \dsm{} as a representation learning framework in the competing risks scenario. We compare its ability to transfer knowledge across multiple competing risks to relevant deep learning alternatives.

\begin{table}[!t]
    \centering
    \caption{Knowledge transfer across tasks and representation learning capability on the \textbf{SYNTHETIC} dataset. Representations were trained on Event 1 and used to predict relative risks for a held-out set on Event 2 using the CPH model.}
    \label{tab:transfer}
    \vspace{0.1in}
    \begin{tabular}{l|c} 
    \toprule \midrule
        Model& C-Index (90\%-CI)\\ \midrule
        \textbf{NNMF} & $0.5940\pm0.0044$\\ 
        \textbf{VAE}& $0.6494\pm0.0044$  \\
        \textbf{K-PCA}&$0.7422\pm 0.0055$\\
         \midrule
        \textbf{DeepSurv}&$0.6988\pm0.0038$\\ 
        \textbf{DeepHit}&$0.7688\pm 0.0040$ \\ \midrule
        \textbf{DSM} & \textbf{0.7724} $\pm$ \textbf{0.0025}\\ \midrule 
        \bottomrule
    \end{tabular}
    \vspace{-0.1in}
\end{table}

We divide the SYNTHETIC data into two equal subsets of 15,000 samples each. For the first set we discard all records that had \textbf{Event 2} before \textbf{Event 1}. For the second set, we perform similar preprocessing and discard all rows where \textbf{Event 1} occurred before \textbf{Event 2}. This effectively converts the two subsets into single event censored datasets for \textbf{Event 1} and \textbf{Event 2} respectively. We train \dsm{}, \textit{DeepSurv} and \textit{DeepHit} on the first half of the dataset for the prediction of \textbf{Event 1}. The learnt model is then used to extract representations for the second subset. The output of the final layer is exploited as an overcomplete representation of the original set of covariates of the individual observation. In both cases, we tune the models by brute-force over one and two hidden layers and dimensionality of the hidden layers in $\{25, 50, 100 \}$. 

For completeness, we also experiment with Kernel-PCA \textbf{(K-PCA)} \cite{scholkopf1997kernel}, Non-Negative Matrix factorization (\textbf{NNMF}) \cite{lee2001algorithms} and modern Variational Auto Encoders (\textbf{VAE}) to learn latent representations. Note that as compared to \textit{DeepSurv} and \dsm{},  \textbf{K-PCA}, \textbf{NNMF} and \textbf{VAE} are intrinsic methods that do not have access to the label of the original risk (\textbf{Event 1}) at training time and hence are limited in their expressive capability. 

Once the representations are extracted for the second subset of the data, a linear Cox Proportional Hazards (CPH) Model is trained on them for the competing risk (\textbf{Event 2}). Table \ref{tab:transfer} presents the result of concordance of the learnt CPH model on the extracted embeddings. DSM outperforms the competing baselines.

\section{Model Complexity and Scalability }

We would like to stress that the advantage of \method{} is not only in terms of competitive predictive performance, but also in the ability to manage computational and inference complexity. 
Since \dsm{} involves making reasonable parametric assumptions, inference requires fewer parameters to learn as compared to the considered alternative approaches. 
In this section, we compare the training time and the model complexity in terms of number of parameters of \dsm{} vis-à-vis the established deep learning baselines, \textbf{DeepHit} and \textbf{DeepSurv}, as well as the linear Cox Proportional Hazards regression \textbf{CPH}. 

\begin{figure}[!t]
    \centering
    \begin{minipage}{0.4\linewidth}
        \includegraphics[width=\linewidth]{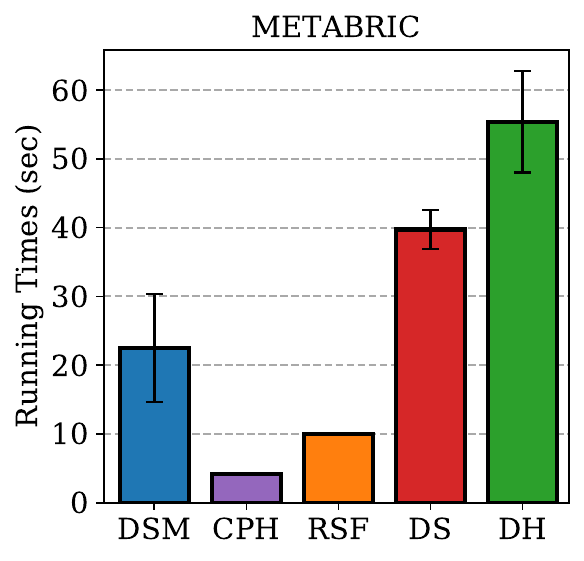}
    \end{minipage}%
    \begin{minipage}{0.4\linewidth}
        \includegraphics[width=\linewidth]{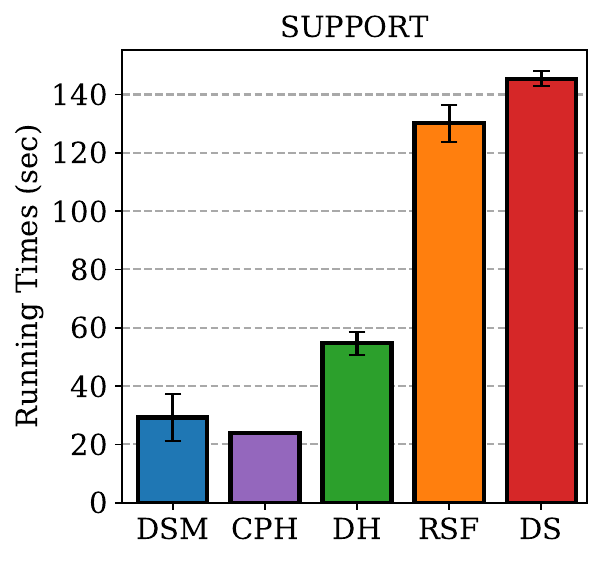}
    \end{minipage}%
    \caption{Comparison of training times required. Parameter inference with \dsm{} is faster than other deep learning approaches, and it scales better with data size.}
    \label{fig:runnintime}
\end{figure}
\begin{figure}[!t]
    \centering
    \vspace{-1em}
    \begin{minipage}{0.4\linewidth}
        \includegraphics[width=\linewidth]{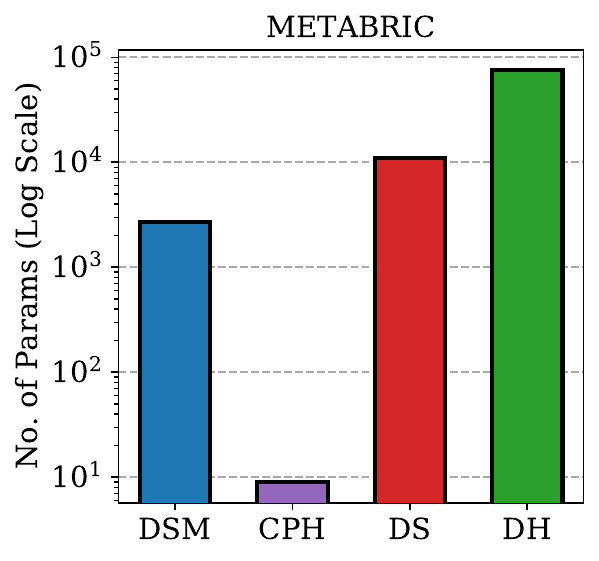}
    \end{minipage}%
    \begin{minipage}{0.4\linewidth}
        \includegraphics[width=\linewidth]{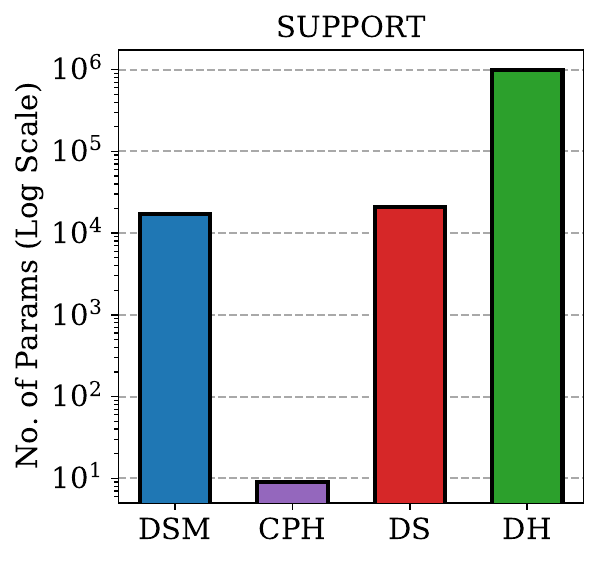}
    \end{minipage}%
    \caption{Number of learnable parameters in best architectures. \dsm{} requires fewer parameters than deep learning alternatives.}
    \label{fig:paramsize}
    \vspace{-1em}
\end{figure}

From Figures \ref{fig:runnintime} and \ref{fig:paramsize}, the advantage of \dsm{} in runtime and space complexity vs.\ considered deep learning alternative models is clearly visible. Note that while RSF is faster in training on METABRIC, it scales poorly with increasing amount of data as evidenced by slower runtime on the larger SUPPORT dataset.
Specifications of the machine used to benchmark performance are provided in Appendix \ref{apx:benchmarking}.

\section{Conclusion and Future Work}

We proposed \method{}, a novel fully-parametric approach to estimate time-to-event in the presence of censoring and competing risks. 
Our approach models the survival function as a weighted mixture of individual parametric survival distributions, and it is trained over a loss function designed to handle both the censored and uncensored data. 
We demonstrated in experiments the benefits of our approach by comparing its performance to other classical and state-of-the-art survival regression approaches on a few diverse datasets, 
and we showed that the representations learnt by deep neural networks in our approach can be leveraged for the knowledge transfer across different competing risks.

Interesting future work directions include extending our approach to multiple censoring scenarios: in this paper we assumed that the data is \textit{right-censored}, but our framework is readily amenable to left truncation and interval censoring as well. 
Additional research can be conducted to further relax parametric assumptions on the survival distributions, and to extend the utility of Deep Survival Machines to allow incorporating multiple sources of data, including complex modalities such as medical imaging.

\section*{Acknowledgements}

This work was partially supported by the Defense Advanced Research Projects Agency under award FA8750-17-2-0130. 

\bibliography{main}
\bibliographystyle{ieeetr}

\newpage

\appendix

\subsection{Loss Function Formulation}
\label{apx:loss}

At test time, \method{} (\dsm{}) describes the survival function of the test individual as a weighted mixture of $K$ survival distribution primitives, and the $K$ weights are a softmax over the output of a neural network. The loss function of \dsm{} is designed to handle both the censored and uncensored data.

\noindent \textbf{Uncensored Loss.} The maximum likelihood estimator for the uncensored data can be written as
\begin{align}
    \nonumber \ln \mathbb{P}(&\mathcal{D}_{U}|  \mathbf{\Theta}) = \ln \bigg(\prod_{i=1}^{\mathcal{|D|}} \mathbb{P} (T=t_i|X=\mathbf{x}_i, \mathbf{\Theta})\bigg)\\
    \nonumber&=\sum_{i=1}^{\mathcal{|D|}}\ln \bigg( \sum_{k=1}^{K} \mathbb{P} (T=t_i|Z, \beta_k, \eta_k)\mathbb{P} (Z | X=\mathbf{x}_i, \bm{w}) \bigg)\\
    \nonumber&=\sum_{i=1}^{|\mathcal{D}|}\ln \bigg( \mathop{\mathbb{E}}\limits_{Z\sim (\cdot| \mathbf{x}_i, \bm{w} )} [\mathbb{P} (T=t_i|Z, \beta_k, \eta_k)] \bigg)\\
    \nonumber&(\text{Applying Jensen's Inequality})\\
    \nonumber&\geq  \sum_{i=1}^{|\mathcal{D}|} \bigg( \mathop{\mathbb{E}}\limits_{Z\sim (\cdot| \mathbf{x}_i, \bm{w} )} [ \ln \mathbb{P} (T=t_i|Z, \beta_k, \eta_k)] \bigg)\\
    \nonumber&=\sum_{i=1}^{|\mathcal{D}|} \bigg(\textsc{Softmax}_{(K)}\big(\ln f(t_i|\beta_{k_i}, \eta_{k_i})\big) \bigg)\\
    \nonumber&\triangleq \textbf{ELBO}_{U}(\Theta)
\end{align}

Here $\mathbf{x}_i$ are the input covariates of the $i$-th observation, and $f(t)$ is the probability density function (PDF) of the primitive distribution. $\beta_{k_i}$ and $\eta_{k_i}$ for the $i$-th observation are parameterized as

$$\beta_{k_i}  = \Tilde{\beta}_k + \texttt{act}(\Phi_\theta(\bm{x}_i)^{\top}\bm{\zeta} ),$$
$$\eta_{k_i} = \Tilde{\eta_k} + \texttt{act}(\Phi_\theta(\bm{x}_i)^{\top}\bm{\xi})$$

where \texttt{act}$(\cdot)$ is the \textsf{SELU} activation function if Weibull is used as the primitive distribution and the \textsf{Tanh} activation function if Log-Normal is used as the primitive distribution. $\mathbf{\Phi}(.)$ is a Multilayer Perceptron. 

\noindent \textbf{Censoring Loss.}
As above, the lower bound of the censored observations can be written as
\begin{align}
\nonumber\ln \mathbb{P}(\mathcal{D}_{C}| \Theta) &= \ln \bigg(\prod_{i=1}^{\mathcal{|D|}} \mathbb{P} (T>t_i|X=\mathbf{x}_i, {\Theta})\bigg) \\
 \nonumber&\geq  \sum_{i=1}^{|\mathcal{D}|} \bigg( \mathop{\mathbb{E}}\limits_{Z\sim (\cdot| \mathbf{x}_i, w )} [ \ln \mathbb{P} (T> t_i|Z, \beta_k, \eta_k)] \bigg)\\
 \nonumber&=\sum_{i=1}^{|\mathcal{D}|} \bigg(\textsc{Softmax}_{(K)}\big(\ln S(t_i|\beta_{k_i}, \eta_{k_i})\big) \bigg)\\
    \nonumber&\triangleq  \textbf{ELBO}_{C}(\Theta)
\end{align}

$S(t)$ is the survival function of the primitive distribution. % The PDF $f(t)$ and survival function $S(t)$ of Weibull and Log-Normal are listed in Table \ref{tab:dists}.

For the scenario of $M$ \textit{competing risks}, $\textbf{ELBO}_{U_m}(\Theta)$ and $\textbf{ELBO}_{C_m}(\Theta)$ are computed for the $m$-th competing risk by treating other events as censoring. The total loss can be written as
$$ \mathcal{L} = \sum_{m=1}^{M} \textbf{ELBO}_{U_m}(\Theta) + \alpha \cdot \textbf{ELBO}_{C_m}(\Theta)  + \mathcal{L}_{\text{prior}_n}   $$

\subsection{Results in Tabular Format}
\label{apx:results}

In this section, we provide the comparison of the performances of \method{} (\dsm{}) with the baseline approaches using $C^{td}$ at different event time horizons. The $C^{td}$ was evaluated at the 25\%, 50\%, 75\% quantiles of event times. The mean and the 90\% confidence interval of the $C^{td}$ were computed using 5-fold cross validation. 

The results of two single-risk datasets, SUPPORT and METABRIC, are respectively shown in Table \ref{tab:support_res} and Table \ref{tab:metabric_res}. To investigate the models' robustness to censoring, we also artificially increased the amount of censoring in training set by censoring a randomly chosen subset which included 25\% or 50\% of the originally uncensored observations in the training data, on both SUPPORT and METABRIC. The results of added censoring are also shown. 

The results of two datasets with competing risks, SYNTHETIC and SEER, are shown respectively in Table \ref{tab:synthetic_res} and Table \ref{tab:seer_res}. cs-CPH and cs-RSF stand for the cause-specific versions of CPH and RSF models.

\begin{table*}[!t]
\centering
\caption{Performance on \textbf{SUPPORT} dataset at different quantiles of event times for different levels of censoring.}
\label{tab:support_res}
\bigskip
\centering
\subfloat[][Default Censoring.]{
\scriptsize

\begin{minipage}[b]{0.4\hsize}
\centering
\begin{tabular}{l|c|c|c}
    \toprule \midrule
    \multirow{3}{*}{Models}&
    \multicolumn{3}{c}{Time-dependent Concordance-Index ($C^{td}$)} \\ [1pt]
    \cline{2-4}
    & \multicolumn{3}{c}{Quantiles of Event Times} \\[1pt] 
    \cline{2-4}
     & $25$\% & $50$\% & $75$\% \\ \midrule
     CPH & $0.794 \pm 0.002$ & $0.756 \pm 0.004$ & $0.733 \pm 0.003$ \\ \midrule
     DeepSurv & $0.804 \pm 0.004$ & $0.767 \pm 0.003$ & $0.746 \pm 0.003$ \\ \midrule
     DeepHit &{$0.822 \pm 0.003$} & {$0.778 \pm 0.002$} & {$0.701 \pm 0.007$} \\ \midrule
     RSF & $0.830 \pm 0.005$ & $0.779 \pm 0.004$ & $0.729 \pm 0.007$ \\ \midrule
     DSM & $0.832 \pm 0.002$ & $0.788 \pm 0.003$ & $0.750 \pm 0.003$ \\ \midrule \bottomrule
\end{tabular}
\end{minipage}
\hspace{1cm}
\begin{minipage}[b]{0.4\hsize}
\centering

\begin{tabular}{l|c|c|c}
    \toprule \midrule
    \multirow{3}{*}{Models}&
    \multicolumn{3}{c}{Brier Score} \\ [1pt]
    \cline{2-4}
    & \multicolumn{3}{c}{Quantiles of Event Times} \\[1pt] 
    \cline{2-4}
     & $25$\% & $50$\% & $75$\% \\ \midrule
     CPH &$0.115\pm0.002$ &$0.171\pm0.002$ &$0.193\pm0.001$ \\ \midrule
     DeepSurv &$0.110\pm0.002$ &$0.165\pm0.002$ &$0.186\pm0.001$ \\ \midrule
     DeepHit & {$0.164 \pm 0.003$} & {$0.285 \pm 0.009$} & {$0.342 \pm 0.020$} \\ \midrule
     RSF &$0.118\pm0.003$ &$0.181\pm0.002$ &$0.203\pm0.002$ \\ \midrule
     DSM &$0.107\pm0.002$ &$0.159\pm0.002$ &$0.188\pm0.002$ \\ \midrule \bottomrule
\end{tabular}
\end{minipage}
}

\subfloat[][25\%+ Censoring.]{
\scriptsize
\begin{minipage}[b]{0.4\hsize}
\centering
\begin{tabular}{l|c|c|c}
    \toprule \midrule
    \multirow{2}{*}{Models}& \multicolumn{3}{c}{Time-dependent Concordance-Index ($C^{td}$)} \\ [1pt]
    \cline{2-4}
    & \multicolumn{3}{c}{Quantiles of Event Times} \\[1pt] 
    \cline{2-4}
     & $25$\% & $50$\% & $75$\% \\ \midrule
     CPH & $0.796 \pm 0.003$ & $0.754 \pm 0.002$ & $0.727 \pm 0.003$ \\ \midrule
     DeepSurv & $0.800 \pm 0.004$ & $0.762 \pm 0.002$ & $0.738 \pm 0.003$ \\ \midrule
     DeepHit & {$0.813 \pm 0.004$} & {$0.770 \pm 0.004$} & {$0.711 \pm 0.008$} \\ \midrule
     RSF & $0.830 \pm 0.004$ & $0.774 \pm 0.001$ & $0.723 \pm 0.005$ \\ \midrule
     DSM & $0.831 \pm 0.002$ & $0.783 \pm 0.003$ & $0.742 \pm 0.003$ \\ \midrule \bottomrule
\end{tabular}
\end{minipage}
\hspace{1cm}
\begin{minipage}[b]{0.4\hsize}
\centering

\begin{tabular}{l|c|c|c}
    \toprule \midrule
    \multirow{3}{*}{Models}&
    \multicolumn{3}{c}{Brier Score} \\ [1pt]
    \cline{2-4}
    & \multicolumn{3}{c}{Quantiles of Event Times} \\[1pt] 
    \cline{2-4}
     & $25$\% & $50$\% & $75$\% \\ \midrule
     CPH &$0.125\pm0.007$ &$0.190\pm0.003$ &$0.224\pm0.001$ \\ \midrule
     DeepSurv &$0.118\pm0.003$ &$0.181\pm0.003$ &$0.213\pm0.001$ \\ \midrule
     DeepHit & {$0.178 \pm 0.004$} & {$0.336 \pm 0.006$} & {$0.436 \pm 0.006$} \\ \midrule
     RSF  &$0.128\pm0.003$ &$0.204\pm 0.003$&$0.243\pm 0.001$ \\ \midrule
     DSM &$0.115\pm0.002$ &$0.176\pm0.003$ &$0.211\pm0.002$ \\ \midrule \bottomrule
\end{tabular}
\end{minipage}
}

\subfloat[][50\%+ Censoring.]{
\scriptsize
\begin{minipage}[b]{0.4\hsize}
\centering
\begin{tabular}{l|c|c|c}
    \toprule \midrule
    \multirow{2}{*}{Models}&
    \multicolumn{3}{c}{Time-dependent Concordance-Index ($C^{td}$)} \\ [1pt]
    \cline{2-4}
    & \multicolumn{3}{c}{Quantiles of Event Times} \\[1pt] 
    \cline{2-4}
     & $25$\% & $50$\% & $75$\% \\ \midrule
     CPH & $0.793 \pm 0.006$ & $0.750 \pm 0.004$ & $0.721 \pm 0.006$ \\ \midrule
     DeepSurv & $0.795 \pm 0.004$ & $0.756 \pm 0.004$ & $0.731 \pm 0.003$ \\ \midrule
     DeepHit & {$0.814 \pm 0.004$} & {$0.771 \pm 0.005$} & {$0.709 \pm 0.006$} \\ \midrule
     RSF & $0.827 \pm 0.002$ & $0.770 \pm 0.003$ & $0.716 \pm 0.005$ \\ \midrule
     DSM & $0.828 \pm 0.002$ & $0.778 \pm 0.004$ & $0.735 \pm 0.004$ \\ \midrule\bottomrule
\end{tabular}
\end{minipage}
\hspace{1cm}
\begin{minipage}[b]{0.4\hsize}
\centering

\begin{tabular}{l|c|c|c}
    \toprule \midrule
    \multirow{3}{*}{Models}&
    \multicolumn{3}{c}{Brier Score} \\ [1pt]
    \cline{2-4}
    & \multicolumn{3}{c}{Quantiles of Event Times} \\[1pt] 
    \cline{2-4}
     & $25$\% & $50$\% & $75$\% \\ \midrule
     CPH &$0.140\pm0.004$ &$0.225\pm0.005$ & $0.278\pm0.005$\\ \midrule
     DeepSurv & $0.131\pm0.004$&$0.210\pm0.004$ & $0.259\pm0.002$ \\ \midrule
     DeepHit & {$0.192 \pm 0.005$} & {$0.365 \pm 0.010$} & {$0.481 \pm 0.017$} \\ \midrule
     RSF &$0.143\pm0.004$ &$0.240\pm0.004$ &$0.301\pm0.002$ \\ \midrule
     DSM &$0.126\pm0.003$ &$0.202\pm0.004$ &$0.244\pm0.002$ \\ \midrule \bottomrule
\end{tabular}
\end{minipage}
}
\end{table*}

\begin{table*}[!t]
\centering
\caption{Performance of \textbf{METABRIC} dataset at different quantiles of event times for different levels of censoring.}
\label{tab:metabric_res}
\bigskip
\centering
\subfloat[][Default Censoring.]{
\scriptsize
\begin{minipage}[b]{0.4\hsize}
\centering
\begin{tabular}{l|c|c|c}
    \toprule \midrule
    \multirow{2}{*}{Models}& \multicolumn{3}{c}{Time-dependent Concordance-Index ($C^{td}$)} \\ [1pt]
    \cline{2-4}
    & \multicolumn{3}{c}{Quantiles of Event Times} \\[1pt] 
    \cline{2-4}
     & $25$\% & $50$\% & $75$\% \\ \midrule
     CPH & $0.620 \pm 0.016$ & $0.620 \pm 0.013$ & $0.629 \pm 0.010$ \\ \midrule
     DeepSurv & $0.634 \pm 0.018$ & $0.635 \pm 0.011$ & $0.637 \pm 0.0010$ \\ \midrule
     DeepHit & {$0.691 \pm 0.016$} & {$0.626 \pm 0.011$} & {$0.585 \pm 0.006$} \\ \midrule
     RSF & $0.713 \pm 0.017$ & $0.673 \pm 0.010$ & $0.644 \pm 0.010$ \\ \midrule
     DSM & $0.720 \pm 0.0116$ & $0.676 \pm 0.009$ & $0.652 \pm 0.009$ \\ \midrule \bottomrule
\end{tabular}
\end{minipage}
\hspace{1cm}
\begin{minipage}[b]{0.4\hsize}
\centering

\begin{tabular}{l|c|c|c}
    \toprule \midrule
    \multirow{3}{*}{Models}&
    \multicolumn{3}{c}{Brier Score} \\ [1pt]
    \cline{2-4}
    & \multicolumn{3}{c}{Quantiles of Event Times} \\[1pt] 
    \cline{2-4}
     & $25$\% & $50$\% & $75$\% \\ \midrule
     CPH &$0.127\pm0.006 $&$0.209\pm0.003 $&$0.249\pm 0.002$\\ \midrule
     DeepSurv &$0.124\pm0.006$ &$0.195\pm0.004$ &$0.226\pm0.007$ \\ \midrule
     DeepHit &{$0.137 \pm 0.003$} & {$0.239 \pm 0.002$} &{$0.284 \pm 0.004$} \\ \midrule
     RSF & $0.119\pm0.006$ & $0.193\pm0.003$& $0.227\pm0.004$ \\ \midrule
     DSM& $0.116\pm0.006$& $0.187\pm0.003$ & $0.222\pm0.005$ \\ \midrule \bottomrule
\end{tabular}
\end{minipage}
}

\subfloat[][25\%+ Censoring.]{
\scriptsize
\begin{minipage}[b]{0.4\hsize}
\centering
\begin{tabular}{l|c|c|c}
    \toprule \midrule
    \multirow{2}{*}{Models}& \multicolumn{3}{c}{Time-dependent Concordance-Index ($C^{td}$)} \\ [1pt]
    \cline{2-4}
    & \multicolumn{3}{c}{Quantiles of Event Times} \\[1pt] 
    \cline{2-4}
     & $25$\% & $50$\% & $75$\% \\ \midrule
     CPH & $0.607 \pm 0.015$ & $0.616 \pm 0.012$ & $0.628 \pm 0.010$ \\ \midrule
     DeepSurv & $0.619 \pm 0.018$ & $0.627 \pm 0.012$ & $0.633 \pm 0.011$ \\ \midrule
     DeepHit & {$0.688 \pm 0.020$} &{$0.618 \pm 0.015$} & {$0.593 \pm 0.002$} \\ \midrule
     RSF & $0.710 \pm 0.016$ & $0.668 \pm 0.009$ & $0.642 \pm 0.010$ \\ \midrule
     DSM & $0.712 \pm 0.010$ & $0.671 \pm 0.010$ & $0.645 \pm 0.010$ \\ \midrule \bottomrule
\end{tabular}
\end{minipage}
\hspace{1cm}
\begin{minipage}[b]{0.4\hsize}
\centering

\begin{tabular}{l|c|c|c}
    \toprule \midrule
    \multirow{3}{*}{Models}&
    \multicolumn{3}{c}{Brier Score} \\ [1pt]
    \cline{2-4}
    & \multicolumn{3}{c}{Quantiles of Event Times} \\[1pt] 
    \cline{2-4}
     & $25$\% & $50$\% & $75$\% \\ \midrule
     CPH &$0.135\pm0.007$ &$0.230\pm0.003$ &$0.288\pm0.003$ \\ \midrule
     DeepSurv &$0.131\pm0.006$ &$0.216\pm0.004$ &$0.260\pm0.007$ \\ \midrule
     DeepHit &{$0.149 \pm 0.003$} &{$0.271 \pm 0.002$} &{$0.335 \pm 0.002$} \\ \midrule
     RSF &$0.127\pm0.007$ &$0.214\pm0.003$ &$0.263\pm0.004$ \\ \midrule
     DSM &$0.125\pm0.007$&$ 0.210\pm0.003 $&$ 0.264\pm0.005$
 \\ \midrule \bottomrule
\end{tabular}
\end{minipage}
}

\subfloat[][50\%+ Censoring.]{
\scriptsize
\begin{minipage}[b]{0.4\hsize}
\begin{tabular}{l|c|c|c}
    \toprule \midrule
    \multirow{2}{*}{Models}& \multicolumn{3}{c}{Time-dependent Concordance-Index ($C^{td}$)} \\ [1pt]
    \cline{2-4}
    & \multicolumn{3}{c}{Quantiles of Event Times} \\[1pt]
    \cline{2-4}
     & $25$\% & $50$\% & $75$\% \\ \midrule
     CPH & $0.603 \pm 0.014$ & $0.613 \pm 0.012$ & $0.630 \pm 0.010$ \\ \midrule
     DeepSurv & $0.617 \pm 0.018$ & $0.612 \pm 0.014$ & $0.622 \pm 0.012$ \\ \midrule
     DeepHit & {$0.666 \pm 0.020$} & {$0.601 \pm 0.011$} & {$0.583 \pm 0.006$} \\ \midrule
     RSF & $0.700 \pm 0.016$ & $0.662 \pm 0.010$ & $0.635 \pm 0.010$ \\ \midrule
     DSM & $0.708 \pm 0.009$ & $0.664 \pm 0.010$ & $0.638 \pm 0.010$ \\ \midrule \bottomrule
\end{tabular}
\end{minipage}
\hspace{1cm}
\begin{minipage}[b]{0.4\hsize}
\centering

\begin{tabular}{l|c|c|c}
    \toprule \midrule
    \multirow{3}{*}{Models}&
    \multicolumn{3}{c}{Brier Score} \\ [1pt]
    \cline{2-4}
    & \multicolumn{3}{c}{Quantiles of Event Times} \\[1pt] 
    \cline{2-4}
     & $25$\% & $50$\% & $75$\% \\ \midrule
     CPH &$0.148\pm0.009$ &$0.264\pm0.004$ &$0.352\pm0.005$ \\ \midrule
     DeepSurv &$0.145\pm0.008$ &$0.251\pm0.006$ &$0.322\pm0.009$ \\ \midrule
     DeepHit &{$0.166 \pm 0.003$} & {$0.321 \pm 0.004$} & {$0.418 \pm 0.010$} \\ \midrule
     RSF &$0.141\pm0.008$ &$0.248\pm0.004$ &$0.324\pm0.006$ \\ \midrule
     DSM &$0.137\pm0.008$ &$0.238\pm0.003$ &$0.305\pm0.007$ \\ \midrule \bottomrule
\end{tabular}
\end{minipage}
}
\end{table*}

\begin{table*}[!ht]
\centering
\caption{$C^{td}$ for competing risks on \textbf{SYNTHETIC}.}
\label{tab:synthetic_res}
\bigskip
\centering
\subfloat[][Event 1.]{
\footnotesize
\begin{tabular}{l|c|c|c|c}
    \toprule \midrule
    \multirow{2}{*}{Models}& \multicolumn{4}{c}{Quantiles of Event Times} \\ \cline{2-5}
     & $25$\% & $50$\% & $75$\% & $100$\%\\ \midrule
     cs-CPH & $0.570 \pm 0.016$ & $0.553 \pm 0.012$ & $0.542 \pm 0.009$ & $0.528 \pm 0.009$ \\ \midrule
     FG & {$0.610 \pm 0.001$} & {$0.587 \pm 0.002$} & {$0.568 \pm 0.003$} & {$0.548 \pm 0.004$} \\ \midrule
     cs-RSF & $0.680 \pm 0.011$ & $0.663 \pm 0.009$ & $0.644 \pm 0.007$ & $0.569 \pm 0.005$ \\ \midrule
     DeepHit & {$0.796 \pm 0.009$} & {$0.765 \pm 0.005$} & {$0.726 \pm 0.005$} & {$0.635 \pm 0.007$} \\ \midrule
     DSM & $0.798 \pm 0.010$ & $0.759 \pm 0.008$ & $0.724 \pm 0.004$ & $0.670 \pm 0.005$ \\ \midrule \bottomrule

\end{tabular}
}

\subfloat[][Event 2.]{
\footnotesize
\begin{tabular}{l|c|c|c|c}
    \toprule \midrule
    \multirow{2}{*}{Models}& \multicolumn{4}{c}{Quantiles of Event Times} \\ \cline{2-5}
     & $25$\% & $50$\% & $75$\% & $100$\%\\ \midrule
     cs-CPH & $0.591 \pm 0.024$ & $0.563 \pm 0.018$ & $0.548 \pm 0.015$ & $0.533 \pm 0.013$ \\ \midrule
     FG & {$0.634 \pm 0.008$} &{$0.595 \pm 0.008$} & {$0.575 \pm 0.007$} & {$0.552 \pm 0.005$} \\ \midrule
     cs-RSF & $0.687 \pm 0.007$ & $0.663 \pm 0.011$ & $0.638 \pm 0.010$ & $0.571 \pm 0.011$ \\ \midrule
     DeepHit & {$0.803 \pm 0.004$} & {$0.761 \pm 0.005$} & {$0.726 \pm 0.004$} & {$0.618 \pm 0.014$} \\ \midrule
     DSM & $0.803 \pm 0.011$ & $0.762 \pm 0.009$ & $0.729 \pm 0.006$ & $0.672 \pm 0.006$ \\ \midrule \bottomrule

\end{tabular}
}
\end{table*}

\begin{table*}[!ht]
\centering
\caption{$C^{td}$ for competing risks on \textbf{SEER}.}
\label{tab:seer_res}
\bigskip
\centering
\subfloat[][Breast Cancer (BC).]{
\footnotesize
\begin{tabular}{l|c|c|c|c}
    \toprule \midrule
    \multirow{2}{*}{Models}& \multicolumn{4}{c}{Quantiles of Event Times} \\ \cline{2-5}
     & $25$\% & $50$\% & $75$\% & $100$\%\\ \midrule
     cs-CPH & $0.891 \pm 0.004$ & $0.849 \pm 0.004$ & $0.827 \pm 0.004$ & $0.807 \pm 0.003$ \\ \midrule
     FG & { $0.876 \pm 0.002$} & {$0.840 \pm 0.002$} & {$0.816 \pm 0.002$} & {$0.798 \pm 0.002$} \\ \midrule
     cs-RSF & $0.897 \pm 0.002$ & $0.852 \pm 0.005$ & $0.823 \pm 0.004$ & $0.801 \pm 0.003$ \\ \midrule
     DeepHit & {$0.899 \pm 0.001$} & {$0.863 \pm 0.003$} & {$0.838 \pm 0.003$} & {$0.815 \pm 0.002$} \\ \midrule
     DSM & $0.904 \pm 0.001$ & $0.861 \pm 0.003$ & $0.840 \pm 0.004$ & $0.820 \pm 0.003$ \\ \midrule \bottomrule

\end{tabular}
}

\subfloat[][Cardiovascular Disease (CVD).]{
\footnotesize
\begin{tabular}{l|c|c|c|c}
    \toprule \midrule
    \multirow{2}{*}{Models}& \multicolumn{4}{c}{Quantiles of Event Times} \\ \cline{2-5}
     & $25$\% & $50$\% & $75$\% & $100$\%\\ \midrule
     cs-CPH & $0.897 \pm 0.007$ & $0.890 \pm 0.006$ & $0.891 \pm 0.002$ & $0.889 \pm 0.002$ \\ \midrule
     FG & {$0.842 \pm 0.008$} &{$0.844 \pm 0.006$} & {$0.854 \pm 0.004$} & {$0.857 \pm 0.004$} \\ \midrule
     cs-RSF & $0.845 \pm 0.006$ & $0.832 \pm 0.008$ & $0.823 \pm 0.008$ & $0.816 \pm 0.009$ \\ \midrule
     DeepHit & {$0.902 \pm 0.003$} & { $0.893 \pm 0.003$} & {$0.893 \pm 0.001$} & {$0.889 \pm 0.001$} \\ \midrule
     DSM & $0.894 \pm 0.004$ & $0.893 \pm 0.004$ & $0.893 \pm 0.001$ & $0.890 \pm 0.001$ \\ \midrule \bottomrule

\end{tabular}
}
\end{table*}

\subsection{Hyperparameter Tuning for the Baselines}
\label{apx:baselines}

We compared the performance of \method{} (\dsm{}) to several competing baseline approaches. In this section, we provide details of the hyperparameter tuning for each baseline approach. The hyperparameters tuned for Random Survival Forests (RSF)~\cite{ishwaran2008random} and \textit{DeepHit}~\cite{lee2018deephit} are described as below, and the best set of hyperparameters was chosen based on the time-dependent Concordance-Index $C^{td}$~\cite{antolini2005time} on the validation set. For Cox Proportional Hazards (CPH) model~\cite{cox1972regression}, we used the default settings in the python \texttt{PySurvival} library.\footnote{\url{https://square.github.io/pysurvival/}} For \textit{DeepSurv}~\cite{katzman2018deepsurv}, We directly used the hyperparameters provided in the \textit{DeepSurv} GitHub repository.\footnote{\url{https://github.com/jaredleekatzman/DeepSurv/tree/master/experiments/deepsurv}} For Fine-Gray (FG) model ~\cite{fine1999proportional}, we used the default settings in the R \texttt{cmprsk} package.\footnote{\url{https://cran.r-project.org/web/packages/cmprsk/cmprsk.pdf}}

\textbf{Random Survival Forests (RSF):} The number of trees in the forest was selected from $[10, 20, 50, 100]$ and the maximum depth of the trees was set to 4.

\textbf{DeepHit (DH):} We followed the experiment settings provided in the \textit{DeepHit} GitHub repository.\footnote{\url{https://github.com/chl8856/DeepHit/}} The number of layers in the shared sub-network and in each cause-specific (CS) sub-network was selected from $[1, 2, 3, 5]$; the number of nodes in each layer was selected from $[50, 100, 200, 300]$; the activation function was selected from [\textsf{RELU}, \textsf{ELU}, \textsf{Tanh}]; and the coefficients $\alpha_k$ for trading off the ranking losses of the $k$ competing risks were chosen from $[0.1, 0.5, 1.0, 3.0, 5.0]$. We generated 10 settings by randomly sampling each hyperparameter from the given lists of candidates 10 times, and selected the best set of hyperparameters which had the highest validation $C^{td}$. The hyperparameters for each dataset are shown in Table \ref{tab:deephit_param}. 

% \multirow{2}{*}{Models}& \multicolumn{3}{c}{Event horizon quantiles} \\ \cline{2-4}

\begin{table*}[!ht]
    \centering
       \caption{The hyperparameters of \textit{DeepHit} for each dataset.}
           \label{tab:deephit_param}
\vspace{0.1in}
    \begin{tabular}{l|l|c|c|c|c|c|c}
        \toprule \midrule
        \multirow{2}{*}{Dataset} & \multirow{2}{*}{Type} & \multicolumn{2}{c|}{Shared Sub-network}  & \multicolumn{2}{c|}{CS Sub-network} & \multirow{2}{*}{Activation} & \multirow{2}{*}{$\alpha$}  \\ \cline{3-6}
         & & No. Layers & No. Nodes & No. Layers & No. Nodes & & \\\midrule
         \textbf{SUPPORT} & Single Risk & $3$ & $100$ & $3$ & $300$ & \textsf{eLU} & $0.1$ \\ \midrule
         \textbf{METABRIC} & Single Risk & $3$ & $100$ & $1$ & $100$ & \textsf{Tanh} & $5.0$ \\ \midrule
         \textbf{SYNTHETIC} & Competing Risks & $3$ & $300$ & $2$ & $50$ & \textsf{eLU} & $0.1$ \\\midrule \textbf{SEER} & Competing Risks & $1$ & $100$ & $2$ & $50$ & \textsf{eLU} & $0.5$ \\\midrule
         \bottomrule
         
    \end{tabular}
 \vspace{-0.1in}
\end{table*}

\subsection{Data Preprocessing}
\label{apx:preprocessing}

Both SUPPORT (single risk) and SEER (competing risks) datasets have missing values. The number and percentage of instances with missing values in each feature of the two datasets are provided in Table \ref{tab:support_missing} and Table \ref{tab:seer_missing}.

$21$ out of the $30$ features in SUPPORT have missing values. For the following $7$ features, we used the suggested normal values\footnote{ \url{http://biostat.mc.vanderbilt.edu/wiki/Main/SupportDesc}} for imputation, which are \textit{Serum Albumin (Day 3)}: $3.5$, \textit{PaO2/(.01*FiO2) (Day 3)}: $333.3$, \textit{Bilirubin (Day 3)}: $1.01$, \textit{Serum Creatinine (Day 3)}: $1.01$, \textit{BUN (Day 3)}: $6.51$, \textit{White Blood Cell Count (Day 3)}: $9$, \textit{Urine Output (Day 3)}: $2,502$, as these values are found to be working well in baseline physiologic data imputation.

$5$ out of the $21$ patient covariates in SEER have missing data. For these $5$ features in SEER, as well as the remaining $14$ features in SUPPORT, we followed the data imputation practice used in \cite{lee2018deephit}: missing data was imputed by the mean for numeric features and the mode for categorical features. We first divided the data into train/validation subsets, and the missing values in each subset were imputed with the mean/mode values of the train set.

\begin{table*}[!ht]
    \centering
       \caption{Statistics of the missing values in each feature of SUPPORT dataset.}
           \label{tab:support_missing}
\vspace{0.1in}
    \begin{tabular}{l|l|l|l|l|l}
        \toprule \midrule
        Feature Name & No. Instances & Feature Name & No. Instances  \\ \hline
        \texttt{Years of Education} & $1,634$ ($17.9\%$)  & \texttt{Income} & $2,982$ ($32.8\%$) \\ \hline 
        \texttt{SUPPORT Coma Score} & $1$ ($0.0\%$)  & \texttt{Average TISS (Days 3-25)} & $82$ ($0.9\%$) \\ \hline \texttt{Race} & $42$ ($0.5\%$)  & \texttt{Mean Arterial Blood Pressure (Day 3)} & $1$ ($0.0\%$) \\ \hline \texttt{White Blood Cell Count (Day 3)} & $212$ ($2.3\%$)  & \texttt{Heart Rate (Day 3)} & $1$ ($0.0\%$) \\ \hline \texttt{Respiration Rate (Day 3)} & $1$ ($0.0\%$)  & \texttt{Temperature (Day 3)} & $1$ ($0.0\%$) \\ \hline \texttt{PaO2/(.01*FiO2) (Day 3)} & $2,325$ ($25.5\%$)  & \texttt{Serum Albumin (Day 3)} & $3,372$ ($37.0\%$) \\ \hline \texttt{Bilirubin (Day 3)} & $2,601$ ($28.6\%$)  & \texttt{Serum Creatinine (Day 3)} & $67$ ($0.7\%$) \\ \hline \texttt{Serum Sodium (Day 3)} & $1$ ($0.0\%$)  & \texttt{Serum pH Arterial (Day 3)} & $2,284$ ($25.1\%$) \\ \hline \texttt{Glucose (Day 3)} & $4,500$ ($49.4\%$)  & \texttt{BUN (Day 3)} & $4,352$ ($47.8\%$) \\ \hline \texttt{Urine Output (Day 3)} & $4,862$ ($53.4\%$)  & \texttt{ADL Patient (Day 3)} & $5,641$ ($62.0\%$) \\ \hline 
        \texttt{ADL Surrogate (Day 3)} & $2,867$ ($31.5\%$)  & & \\
        \hline 
         \bottomrule
    \end{tabular}
 \vspace{-0.1in}
\end{table*}

\begin{table*}[!ht]
    \centering
       \caption{Statistics of the missing values in each feature of SEER dataset.}
           \label{tab:seer_missing}
\vspace{0.1in}

    \begin{tabular}{l|l|l|l|l|l}
        \toprule \hline
        Feature Name & No. Instances & Feature Name & No. Instances   \\ \hline
        \texttt{Surgery Type} & $36,524$ ($55.8\%$) & \texttt{Surgery-Beyond Primary Site} & $59,295$ ($90.6\%$) \\ \hline
        \texttt{Surgery-Primary Site} & $28,957$ ($44.2\%$) & \texttt{\makecell{Surgery-Distant Lymph Nodes/ Other Tissues}} & $35,143$ ($53.7\%$)  \\ 
        \hline
        \texttt{No. Lymph Nodes Examined} & $35,143$ ($53.7\%$) & &  \\ \hline
         \bottomrule
    \end{tabular}
 \vspace{-0.1in}
\end{table*}

\subsection{Benchmarking Machine Specifications}
\label{apx:benchmarking}

All experiments except the experiments for \textit{DeepHit} were run on a Linux version 3.10.0-1062.9.1.el7.x86\_64 machine with an Intel(R) Core(TM) i7-3770 CPU @ 3.40GHz (8-core CPU) and RAM 32 GB. The experiments for \textit{DeepHit} were run on a TITAN X (Pascal) GPU cluster (1 GPU) with an Intel(R) Xeon(R) CPU E5-2620 v4 @ 2.10GHz (32-core CPU), NVIDIA driver version 418.74 and CUDA 10.1.

\end{document}